%% file: main.tex
\documentclass[conference,10pt,letterpaper]{IEEEtran}
\input{preamble}

\definecolor{OliveGreen}{rgb}{0,0.6,0}
\providecommand{\keywords}[1]{\textbf{\textit{Key Words:}} #1}
\IEEEoverridecommandlockouts
\begin{document}
\title{Measurement-driven Security Analysis of Imperceptible Impersonation Attacks\thanks{accepted and appears in ICCCN 2020}}

\author{Shasha Li, Karim Khalil,  Rameswar Panda, Chengyu Song \\ Srikanth V. Krishnamurthy , Amit
  K. Roy-Chowdhury, $^*$Ananthram Swami\\ University of California
  Riverside, $^*$US Army Research Laboratory\\
\{sli057,karimk,rpanda002\}@ucr.edu, \{csong,krish\}@cs.ucr.edu,\\ 
{amitrc@ee.ucr.edu}, {ananthram.swami.civ@mail.mil}}

\maketitle

\input{abs1}
\input{intro2}

\input{rel2}

\input{attack2}
\input{exp3}
\input{conc2}

\bibliographystyle{acm}
\bibliography{refs}

\end{document}

%% file: preamble.tex
\usepackage[keeplastbox]{flushend}
\usepackage{xspace} 
\usepackage{graphicx}
\usepackage{amsmath}
\usepackage{amssymb}
\usepackage{cite}
\usepackage{balance}
\usepackage{epsfig}
\usepackage{color}
\usepackage{psfrag}
\usepackage{dsfont}
\usepackage{algpseudocode}
\usepackage{algorithm}
\usepackage{flushend}
\usepackage{multirow}
\usepackage{booktabs}

\newfont{\bbb}{msbm10 scaled 500}

\newfont{\bb}{msbm10 scaled 1100}

\newcommand{\rv}{{\bf r}}

\newcommand{\xv}{{\bf x}}

\newcommand{\zerov}{{\bf 0}}

\newcommand{\Cc}{{\cal C}}

\newcommand{\Sc}{{\cal S}}

\newcommand{\Xc}{{\cal X}}

\newcommand{\argmin}{\operatornamewithlimits{arg\,min}}

\RequirePackage[normalem]{ulem} 
\RequirePackage{color}\definecolor{RED}{rgb}{1,0,0}
\definecolor{BLUE}{rgb}{0,0,1} 

\usepackage{times}
\usepackage{epsf}
\usepackage{epsfig}
\usepackage{mathrsfs}
\usepackage[font=small,labelfont=bf]{caption}
\usepackage{subcaption}
\usepackage{graphics}
\usepackage{graphicx}
\usepackage{epstopdf}
\usepackage{amsmath}
\usepackage{amsfonts}
\usepackage{textcomp}
\usepackage{pdflscape}
\usepackage{afterpage}
\usepackage{color}
\usepackage{comment}
\usepackage{tabularx}
\usepackage{amssymb}
\usepackage{color,soul}
\usepackage{ifpdf}

\newcommand{\squishlist}{
   \begin{list}{$\bullet$}
    { \setlength{\itemsep}{0pt}      \setlength{\parsep}{3pt}
      \setlength{\topsep}{3pt}       \setlength{\partopsep}{0pt}
      \setlength{\leftmargin}{1.3em} \setlength{\labelwidth}{1em}
      \setlength{\labelsep}{0.5em} } }

\newcommand{\squishlisttwo}{
   \begin{list}{$\bullet$}
    { \setlength{\itemsep}{0pt}    \setlength{\parsep}{0pt}
      \setlength{\topsep}{0pt}     \setlength{\partopsep}{0pt}
      \setlength{\leftmargin}{2em} \setlength{\labelwidth}{1.5em}
      \setlength{\labelsep}{0.5em} } }

\newcommand{\squishend}{
    \end{list}  }

\usepackage[hyphens]{url}
\usepackage[hidelinks]{hyperref}
\hypersetup{breaklinks=true}
\usepackage{cite}

%% file: abs1.tex
\begin{abstract}
  The emergence of Internet of Things (IoT) brings about new security challenges
  at the intersection of cyber and physical spaces. One prime example is the
  vulnerability of Face Recognition (FR) based access control in IoT
  systems. While previous research has shown that Deep Neural Network
  (DNN)-based FR systems (FRS) are potentially susceptible to imperceptible
  impersonation attacks, the potency of such attacks in a wide set of scenarios
  has not been throughly investigated. In this paper, we present the first
  systematic, wide-ranging measurement study of the exploitability of DNN-based
  FR systems using a large scale dataset. We find that arbitrary impersonation
  attacks, wherein an arbitrary attacker impersonates an arbitrary target, are
  hard if imperceptibility is an auxiliary goal. Specifically, we show that
  factors such as skin color, gender, and age, impact the ability to carry out
  an attack on a specific target victim, to different extents. We also study the
  feasibility of constructing universal attacks that are robust to different
  poses or views of the attacker's face. Our results show that finding a universal
  perturbation is a much harder problem from the attacker's
  perspective. Finally, we find that the perturbed images do not generalize well
  across different DNN models. This suggests security countermeasures that can
  dramatically reduce the exploitability of DNN-based FR systems.

\keywords{face recognition, imperceptible adversarial perturbation, Internet of Things}

\end{abstract}

%% file: intro2.tex
\section{Introduction}
\label{sec:introduction}


Face-recognition-based biometric authentication has become very
popular in Internet of Things (IoT)~\cite{yang2013intelligent,pentland2000face, manjunatha2017home}.
In fact, according to the International Biometric Group (IBG), face is the second most widely deployed biometric in terms of market share, right after fingerprints~\cite{international2007biometrics}. The most noteworthy applications using face recognition include opening doors~\cite{ibrahim2011study}, activating personalized services by automated identification of users, e.g., smart TV program selector or pervasive software such as Microsoft's Kinect~\cite{zuo2005real,manjunatha2017home}.

Face Recognition Systems (FRSs) are typically trained on known faces, and use the trained model to classify test cases (i.e., when a human presents herself to a camera). The deep learning paradigm has seen significant proliferation in FRSs due to its ability to provide high recognition accuracy~\cite{taigman2014deepface,schroff2015facenet,parkhi2015deep}.

\begin{figure*}[t]
\centering
\includegraphics[width=2.0\columnwidth]{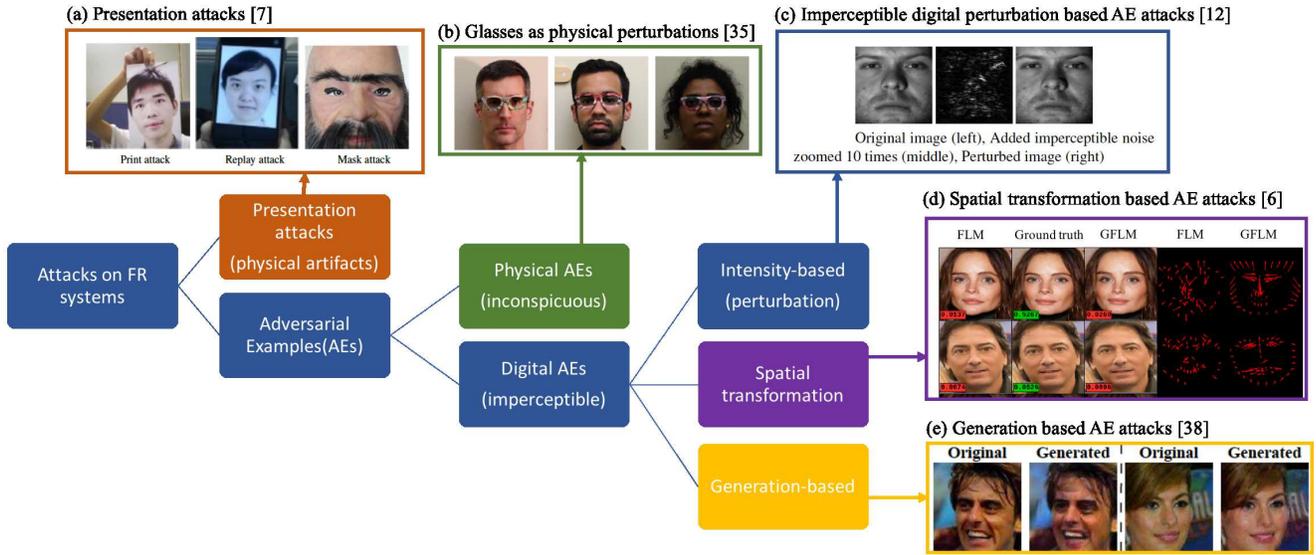}
\caption{Various attacks on Face Recognition Systems. We focus on intensity-based AE attacks in our analysis since they are the kind of attacks explored the most in the literature. Intensity-based AE attacks are fast to carry out and are proven to have high attack success rates.}   
\label{fig:attacks}
\end{figure*}

Due to the ubiquity of FRSs in security-critical applications, their security and reliability have drawn attention and various attacks have been showcased. Early presentation attacks\cite{anjos2011counter,erdogmus2014spoofing,xu2016virtual}
 impersonate a victim's identity by presenting a fake face to FRSs, which could be in the form of photographs, replayed videos, 3D masks etc., as shown in Fig.~\ref{fig:attacks}(a). It has been recently shown that Deep Neural Networks (DNNs) are vulnerable to adversarial examples~\cite{szegedy2013intriguing,goodfellow2014explaining,kurakin2016adversarial}. Adversarial examples are generated in such a manner that humans cannot notice adversarially induced perturbations and correctly classify the images, but the perturbations cause FRSs to misclassify them. Many attack methods~\cite{goswami2018unravelling,dabouei2019fast,song2018attacks,deb2019advfaces} have been proposed to generate adversarial examples for impersonation attacks, among which, intensity-based adversarial examples (Fig.~\ref{fig:attacks}(c)) can be quickly generated and are effective against a variety of FRSs~\cite{goodfellow2014explaining,kurakin2016adversarial}. Intensity-based impersonation attacks add imperceptible perturbation to the original face images such that the FRSs misclassify the perturbed face images (adversarial examples) to be that of the victim. 

While we defer a detailed discussion of related work to \S~\ref{sec:related},
we find that none of previous efforts perform an in depth study on the scope and
effectiveness of such intensity-based impersonation attacks (referred to as impersonation attacks from hereon).
In other words, there seems to be no answer yet to the question ``Can an arbitrary attacker impersonate an
arbitrary victim easily?" The key term here is \emph{easily}. Specifically, if an
attacker were able to add arbitrary amount of perturbations to her own image, 
she certainly could impersonate any victim. However, this would cause the attacker to be stand out,
i.e., her actions could be perceived by observers as strange or even suspicious.
Thus, the perturbation has to be imperceptible---the perturbation used
must be small and inconspicuous. The question that is of interest therefore
becomes "Can the perturbations be kept small in general settings?".

Towards answering this question, we undertake an in depth, 
systematic measurement study of the exploitability of DNN-based FRSs,
using a very large scale dataset of about 2.6 million images.
Our measurement study demonstrates that several factors influence the
imperceptibility of impersonation attacks. We also find that it is more
difficult to fool systems if the attacker has to account for the variability in
her pose/orientation and other environmental conditions such as lighting, or use
the perturbations generated from one DNN model to attack a different model.
Based on the measurements, we suggest security countermeasures that could
significantly enhance the security of FR based IoT access control. In brief, our
contributions in this paper are :

\squishlist
\item We perform an extensive measurement study which shows that the
  efficacy/imperceptibility of impersonation attacks depend on several factors
  such as gender, skin color and age. We quantify the extent to which each of
  these factors affect the attack.

\item We perform an in-depth measurement study to understand the feasibility of
  constructing universal perturbations that make the attack robust to different
  poses or facial orientations of the attacker. We find that this is much harder
  in practice from the attacker's perspective.

\item We show that the use of multiple DNNs for performing FR (check faces
  across DNN models) can render imperceptible impersonation attacks almost
  infeasible.

\squishend

%% file: rel2.tex
\section{Related Work}
\label{sec:related}
\subsection{DNNs based FRSs}
A lot of efforts have targeted the design of highly accurate FRSs. Traditional methods applied hand-crafted features like edges and texture descriptors~\cite{park2010age,li2015face,li2015adaptive,ding2016comprehensive}, which have been used for a long time. Due to the convenience of obtaining large training data and the availability of inexpensive computing power and memory, the trend towards replacing the traditional methods by deep learning methods is increasing. Deep Convolutional Neural Networks (DCNNs) can automatically extract high level representative features from large datasets and have been shown to be invariant to illumination variations, brightness variations, age variations and/or facial orientation~\cite{arsenovic2017facetime}. 
Today the state-of-the-art FR
algorithms are almost all based on end-to-end DCNNs~\cite{taigman2014deepface,schroff2015facenet,parkhi2015deep,sun2015deeply,liu2015targeting}. We use VGG-Face~\cite{parkhi2015deep} in our analysis. VGG-Face is a 39-layer DCNN, and is one of the most well-known and highly accurate face recognition systems.  

\subsection{Presentation Attacks}
It is generally
believed that DNN-based FRSs have extremely high recognition accuracy,
even better than humans. However, this is based on the implicit assumption that
attackers do not actively attempt to fool the system. Recently however, there
have been extensive efforts reported in the literature on attacks targeting 
FRSs 
\cite{duc2009your,findling2012towards,erdogmus2014spoofing,xu2016virtual,goel2018smartbox}.

Many early approaches used by attackers to spoof a FRS, are based on using
fake target faces, which is termed \textit{presentation attack}. In general, attackers hold a non-real face of a target person
in front of the camera to evade the FRS. The attackers could use
photographs\cite{li2014understanding,anjos2011counter}, replayed videos
\cite{chingovska2012effectiveness,zhang2012face}, dummy faces (such as 3D masks)
\cite{erdogmus2014spoofing,kose2013vulnerability}, or 3D virtual reality facial
models displayed on a screen \cite{xu2016virtual} as shown in Fig.~\ref{fig:attacks}(a). While these methods are shown
to successfully lead to attacker misclassification as the target identities,
such attacks, they however require the attacker to overtly indulge in action that
may seem strange or even suspicious to nearby observers.

\begin{figure*}[t]
    \centering
    \begin{subfigure}[t]{0.13\textwidth}
        \centering
        \includegraphics[height=0.62in]{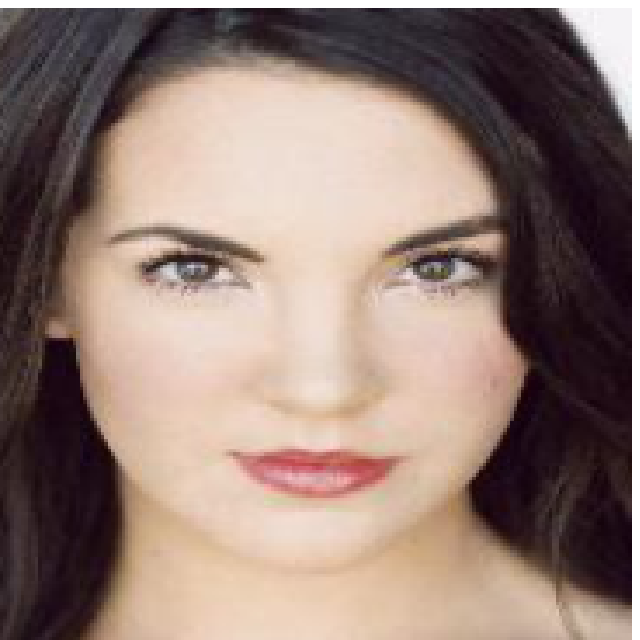}
        \caption{original image}
    \end{subfigure}%
    ~
    \begin{subfigure}[t]{0.21\textwidth}
        \centering
        \includegraphics[height=0.62in]{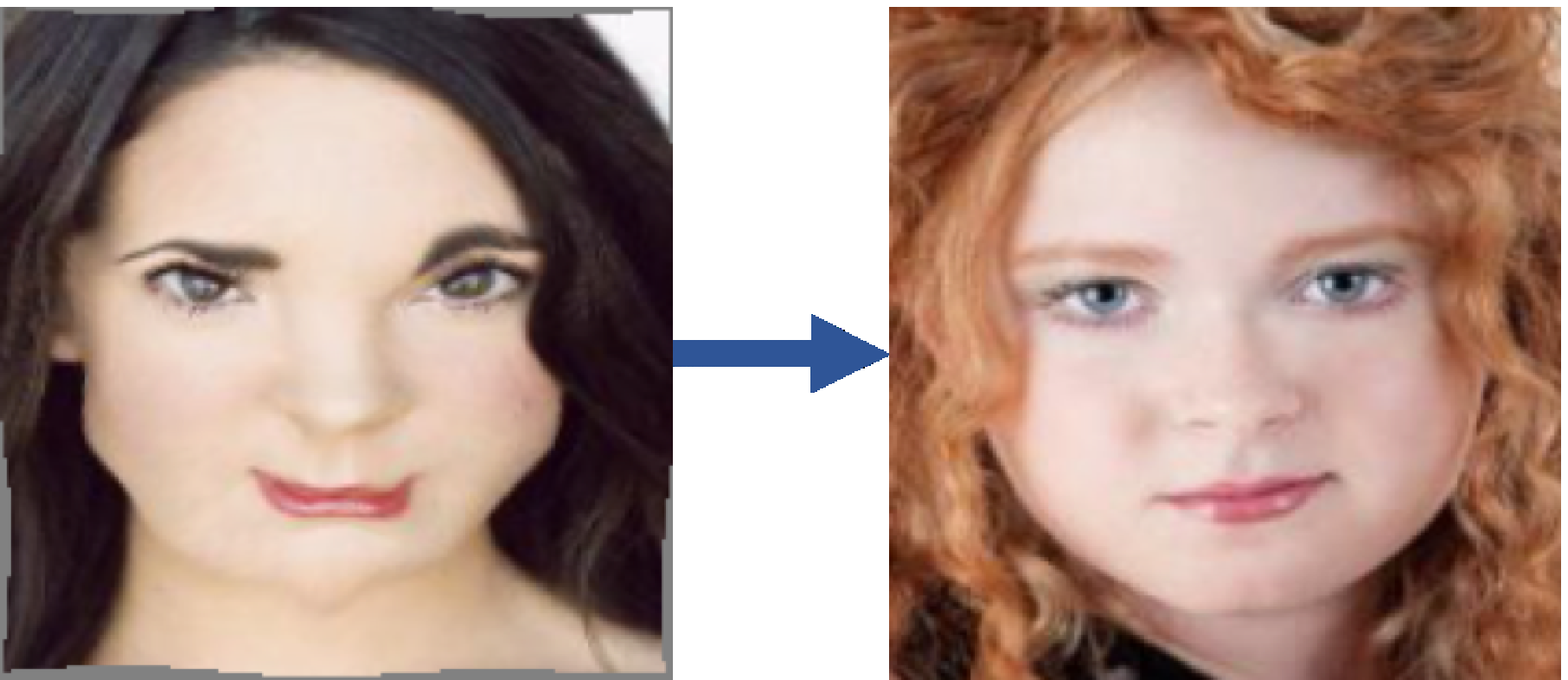}
        \caption{impersonation attack}
    \end{subfigure}%
    ~
    \begin{subfigure}[t]{0.21\textwidth}
        \centering
        \includegraphics[height=0.62in]{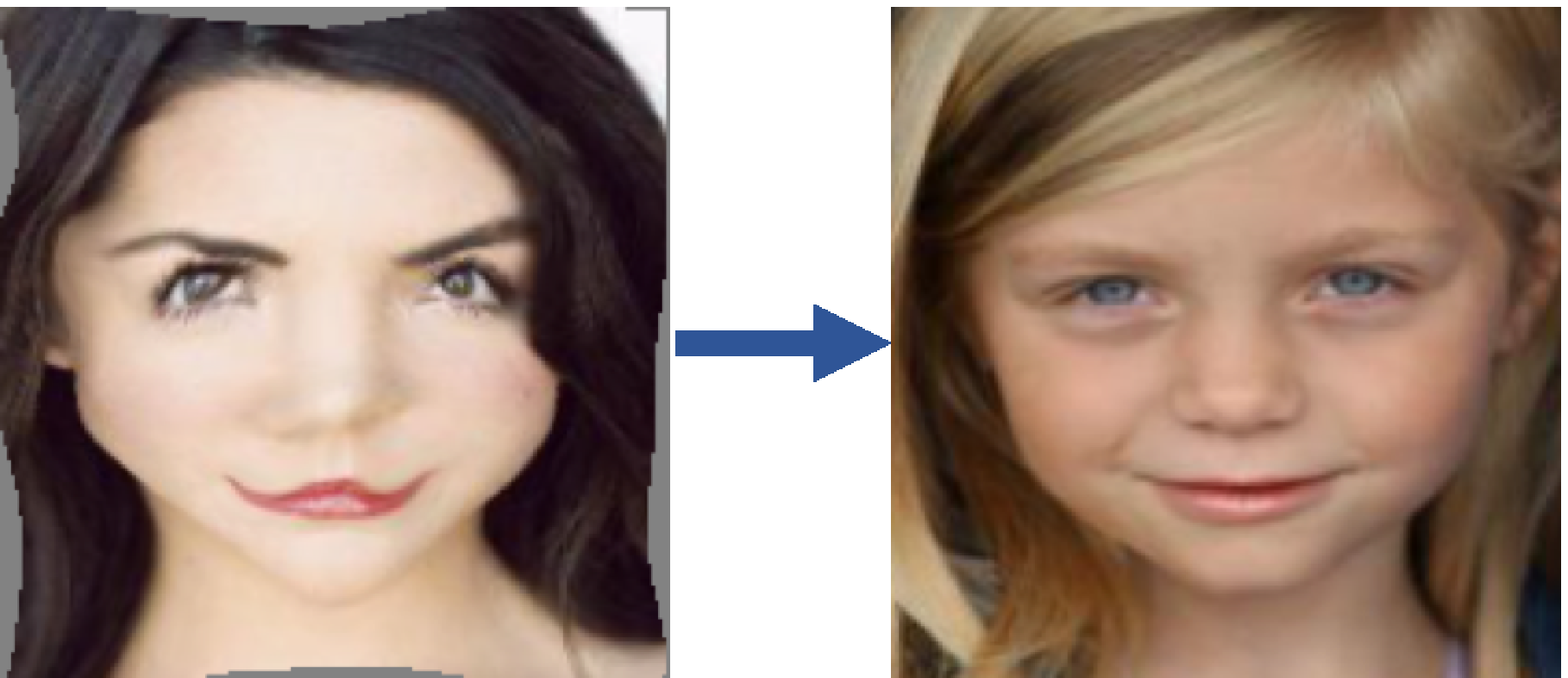}
        \caption{impersonation attack}
    \end{subfigure}%
    ~
    \begin{subfigure}[t]{0.21\textwidth}
        \centering
        \includegraphics[height=0.62in]{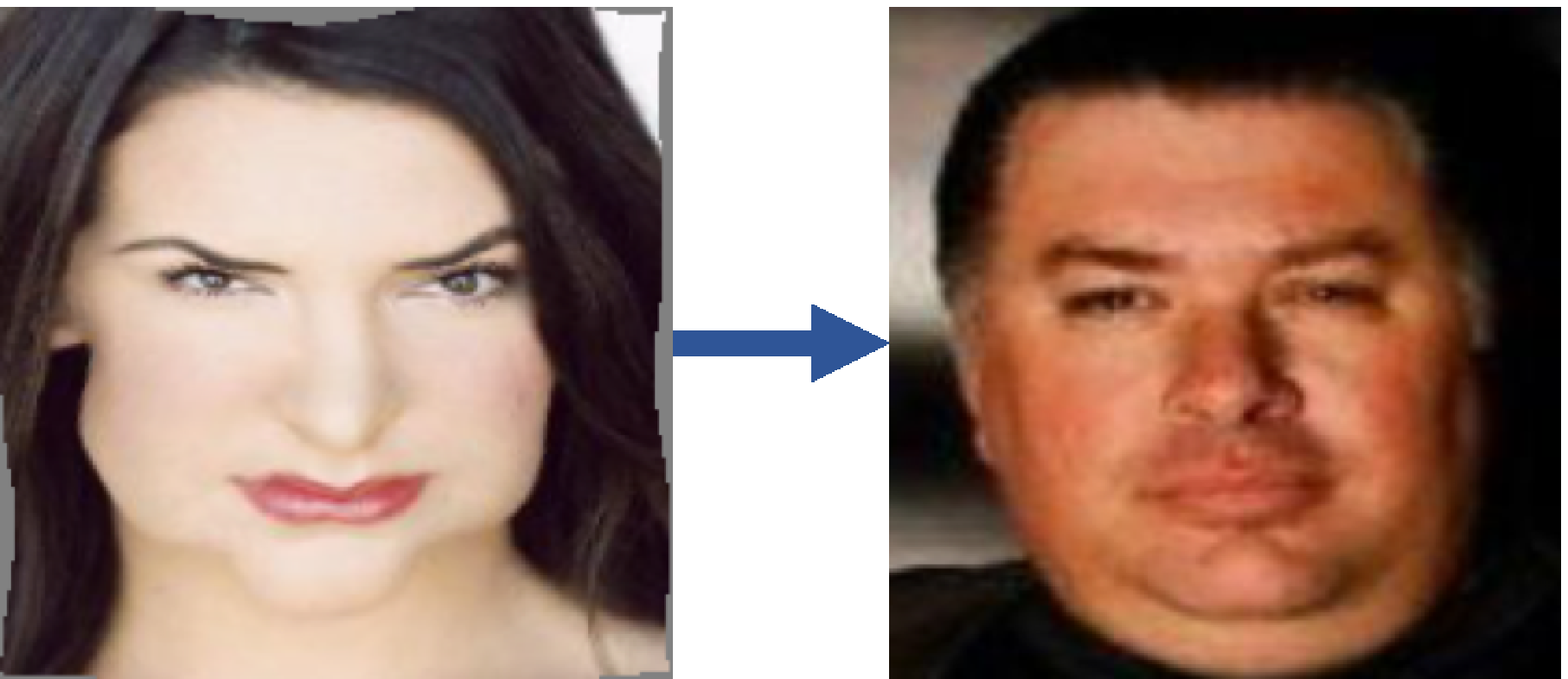}
        \caption{impersonation attack}
    \end{subfigure}%
    ~ 
   \begin{subfigure}[t]{0.21\textwidth}
        \centering
        \includegraphics[height=0.62in]{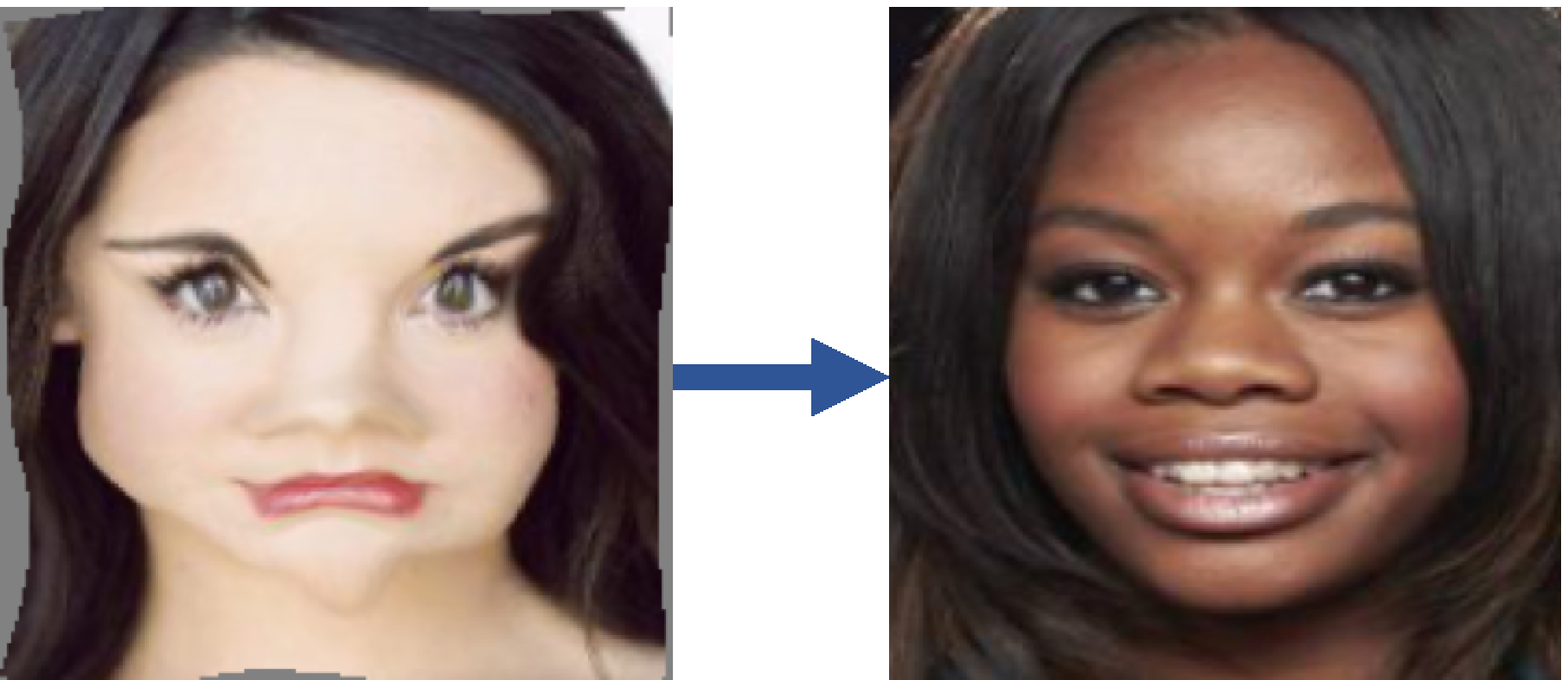}
        \caption{impersonation attack}
    \end{subfigure}%
    \caption{Impersonation attacks using the Fast Landmark Manipulation (FLM) method proposed in~\cite{dabouei2019fast}. (a) shows the original image; (b)-(e) show four impersonation attacks, within each the left image is the adversarial example and the target identity is shown in the right image.} \label{fig:spatial}
\end{figure*}

\subsection{Adversarial Examples for FRSs}
More recently, general DNN-based classifiers \cite{szegedy2013intriguing,li2019stealthy,zhu2020a4} 
have been shown to be vulnerable to adversarial example attacks. Adversarial Examples (AEs) refer to perturbed inputs, which are correctly classified by humans, but misclassified by machine learning systems. In \cite{sharif2016accessorize,sharif2019general}, the authors demonstrate the potential of using adversarial examples to conduct real face attacks on FRSs, i.e., the attackers use their own faces to mount attacks. By wearing special glasses (physical perturbations), the attacker's face can be misclassified by the DDN as shown in Fig.~\ref{fig:attacks}(b). 

In addition to physical AE attacks , various digital AE attack approaches have been proposed, which can be categorized into three kinds as follows.
\begin{itemize}
	\item \textit{Intensity-based AE attacks.} Imperceptible Perturbations are added to the images to change the intensity of each pixel as shown in Fig.~\ref{fig:attacks}(c). \cite{goodfellow2014explaining} hypothesizes that DNNs are vulnerable to AE attacks because of their linear nature and thus proposed the fast gradient sign method (FGSM) for efficiently generating perturbations. \cite{kurakin2016adversarial} extends the FGSM method by applying it multiple times with a small step size. \cite{moosavi2016deepfool} uses a norm minimization based formulation, termed DeepFool, to search for adversarial perturbations by casting it as an optimization problem. \cite{carlini2017towards} introduces new gradient based attack algorithms that are more effective in terms of the adversarial success rates. We use~\cite{kurakin2016adversarial} in our analysis since it can generate adversarial perturbations very fast, which is the key requirement for large-scale analysis (needed to generate these perturbations), and at the same time, it achieves very high attack success rates compared to other fast methods.
	\item \textit{Spatial transformation based AE attacks.} As opposed to manipulating the pixel values, perturbations generated through spatial transformation could result in large $L_{p}$ distance measures, but are perceptually realistic as shown in Fig.~\ref{fig:attacks}(d). \cite{xiao2018spatially} estimates the displacement field for all pixel locations in the input images. \cite{dabouei2019fast} first detects key landmarks of the faces and the displacement field is only defined for the key landmarks.
	\item \textit{Generation-based AE attacks.} \cite{song2018attacks} utilize generative models to generate fake face images as shown in Fig.~\ref{fig:attacks}(e), which are visually similar to the original face images, thus hard to 
cause noticability; at the same time, these have similar feature representations as the target faces, and are thus recognized as the target individuals.
\end{itemize}

There are two different kinds of attack goals viz.: 
\begin{itemize}
	\item Dodging, where the attacker seeks to have one face misidentified as any other different face.
	\item Impersonation, where the attacker seeks to have one face classified as a specific target victim's face, which is harder than the dodging attacks.
\end{itemize}
While dodging attacks are of interest in evading surveillance, impersonation attacks, which are much more targeted, are of more relevance to IoT security. Attackers can leverage this method to gain unauthorized entry, for instance, by bypassing a smart locking mechanism. 
Our work thus focuses on impersonation attacks. The spatial transformation based attacks, that is, Fast Landmark Manipulation Method (FLM) and Grouped Fast Landmark Manipulation Methods (GFLM), are proposed for realizing dodging attacks. We extend these two methods to the impersonation attack. We observe that FLM gives largely deformed facial images as shown in Fig.~\ref{fig:spatial}, which is not imperceptible at all. GFLM, which aims to generate more natural adversarial examples, fails in all the four impersonation attacks. Therefore, it is evident that these types of attacks are not appropriate for impersonation and thus, we do not 
perform additional measurements on such spatial transformation based attack methods. 

We focus on intensity-based AE attacks in our analysis since they are the kind of attacks explored the most in the literature. Intensity-based AE attacks are fast to carry out and have been proven to have extremely high attack success rates. 
Unlike prior works which simply showcase the possibility of such attacks, we do extensive measurements to provide a detailed view of the potency of such attacks in various scenarios and unearth various factors that affect this potency.

%% file: attack2.tex
\section{Imperceptible Impersonation Attack}
\label{sec:attack}
\begin{figure*}[t]
    \centering
    \begin{subfigure}{0.275\columnwidth}
        \centering
        \includegraphics[height=0.8in]{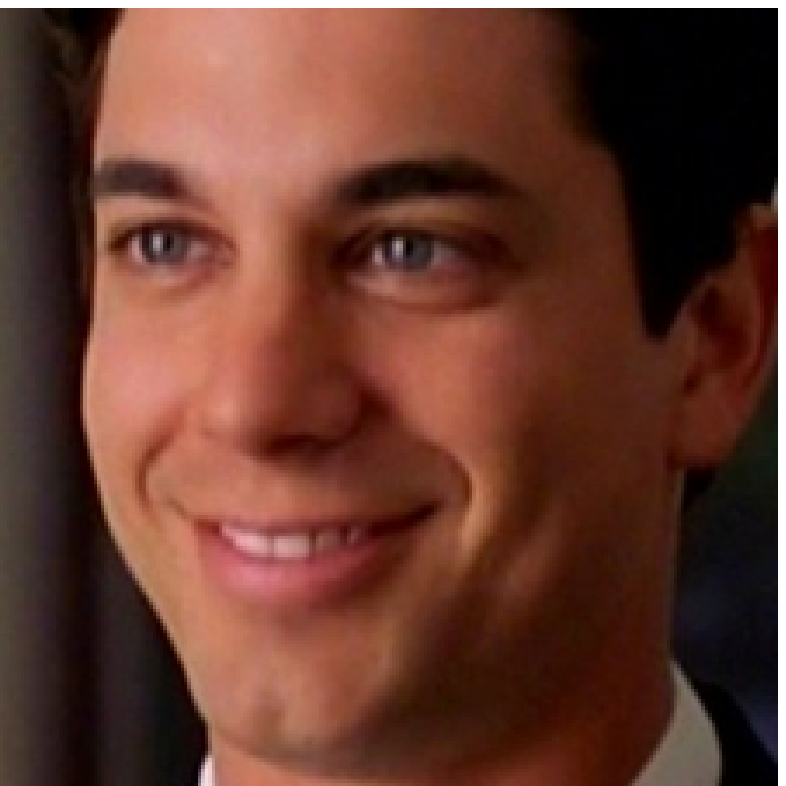}
        \caption{$\sigma=0$}
    \end{subfigure}%
    ~ 
    \begin{subfigure}{0.275\columnwidth}
        \centering
        \includegraphics[height=0.8in]{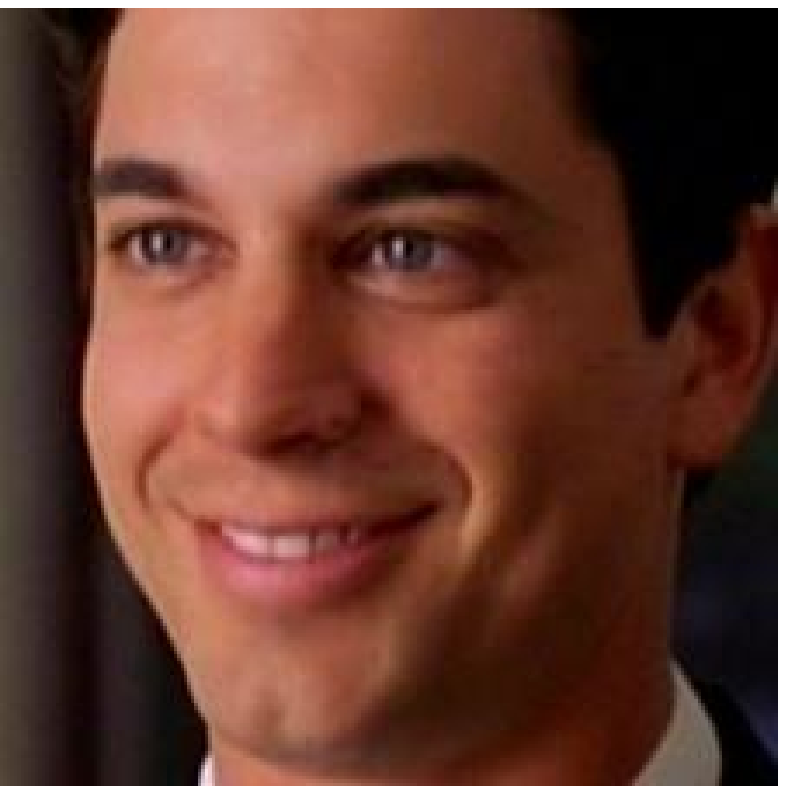}
        \caption{$\sigma=0.8277$}
    \end{subfigure}%
    ~
    \begin{subfigure}{0.275\columnwidth}
        \centering
        \includegraphics[height=0.8in]{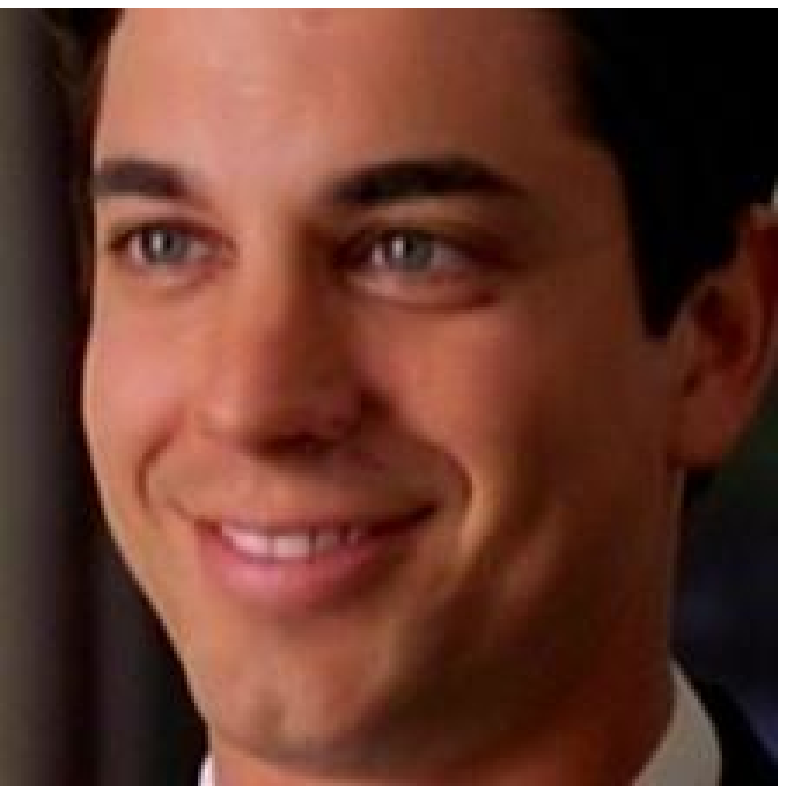}
        \caption{$\sigma=1.0423$}
    \end{subfigure}%
    ~
 \begin{subfigure}{0.275\columnwidth}
        \centering
        \includegraphics[height=0.8in]{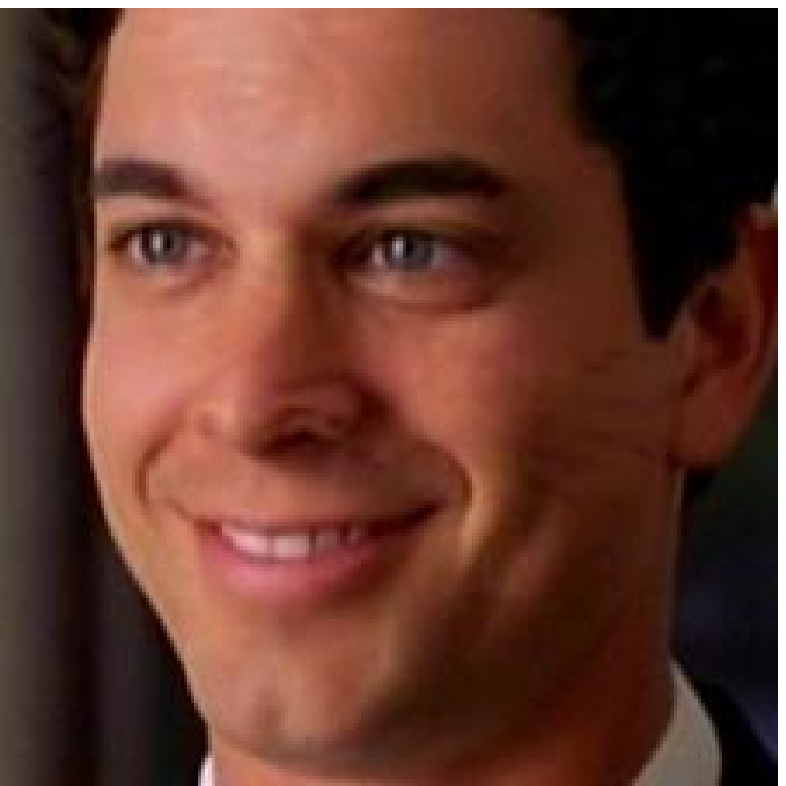}
        \caption{$\sigma=1.5407$}
    \end{subfigure}%
    ~ 
\begin{subfigure}{0.275\columnwidth}
        \centering
        \includegraphics[height=0.8in]{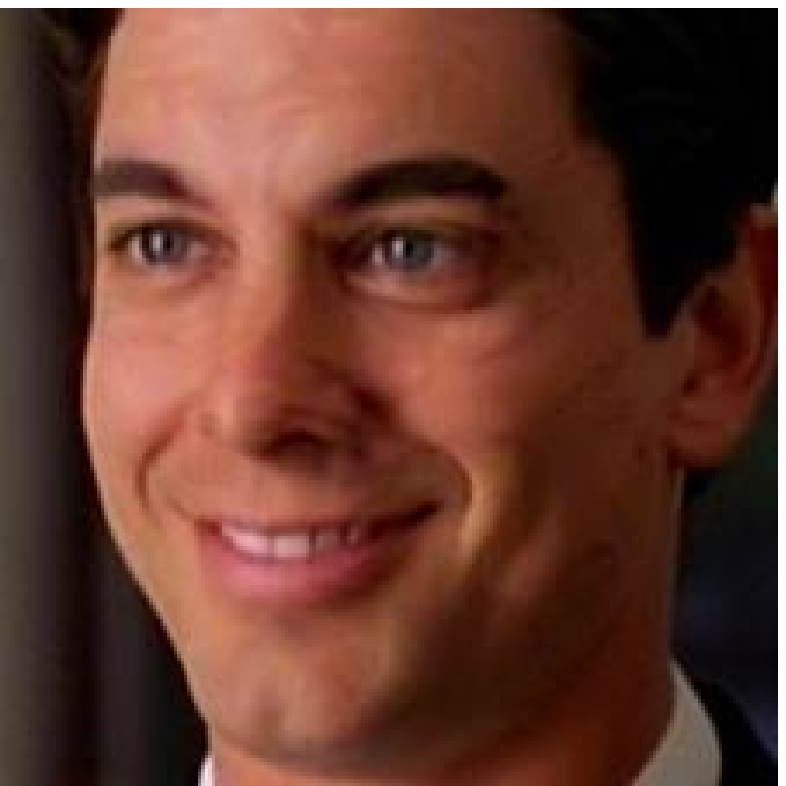}
        \caption{$\sigma=2.0735$}
    \end{subfigure}%
    ~ 
\begin{subfigure}{0.275\columnwidth}
        \centering
        \includegraphics[height=0.8in]{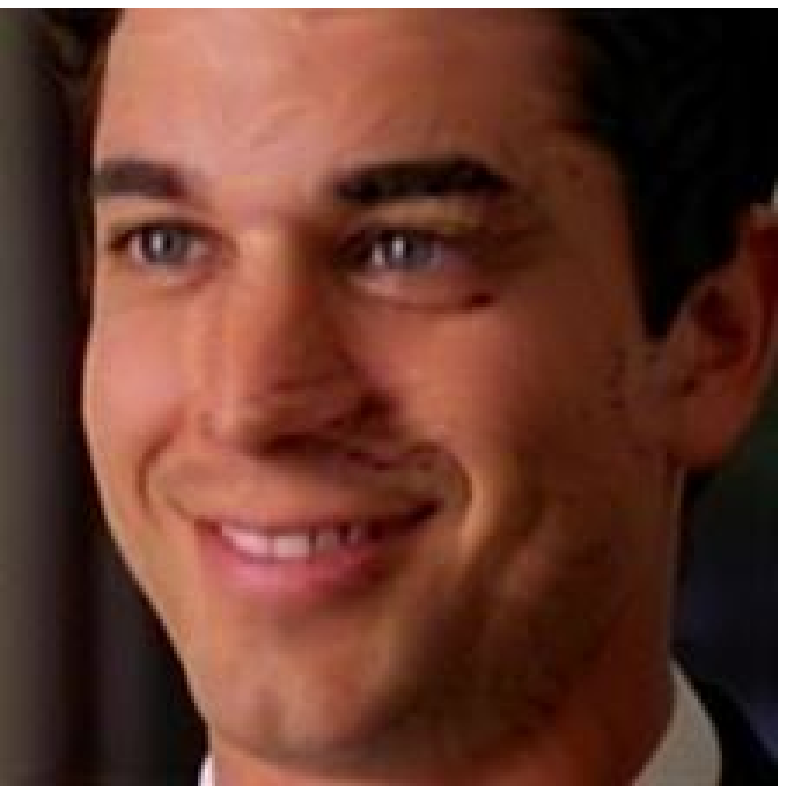}
        \caption{$\sigma=2.7160$}
    \end{subfigure}%
    ~     
\begin{subfigure}{0.275\columnwidth}
        \centering
        \includegraphics[height=0.8in]{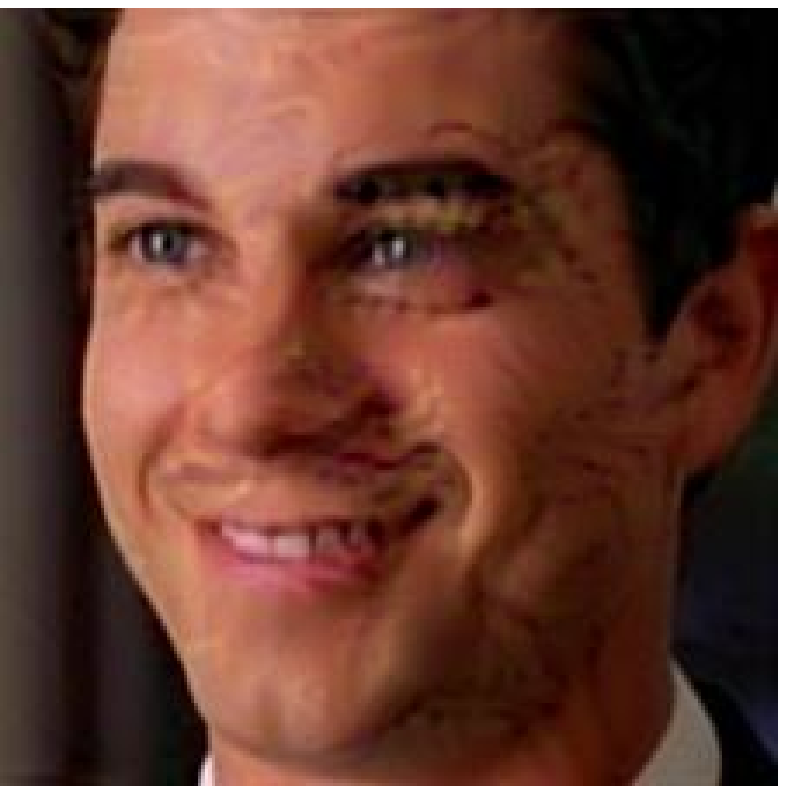}
        \caption{$\sigma=5.00$}
    \end{subfigure}%
    \caption{Perturbed images for different levels of perturbation $\sigma$. In
      (g) with large value of $\sigma=2.7160$, patterns are visible on the forehead,
      left cheek and nose. (The patterns are more visible in color version.)}
\label{fig:perturbedImg}
\end{figure*}

\begin{figure}
    \centering
    \begin{subfigure}{0.275\columnwidth}
        \centering
        \includegraphics[height=0.8in]{./Figs/noise2_7160.eps}
        \caption{}
    \end{subfigure}%
    ~ 
 \begin{subfigure}{0.275\columnwidth}
        \centering
        \includegraphics[height=0.8in]{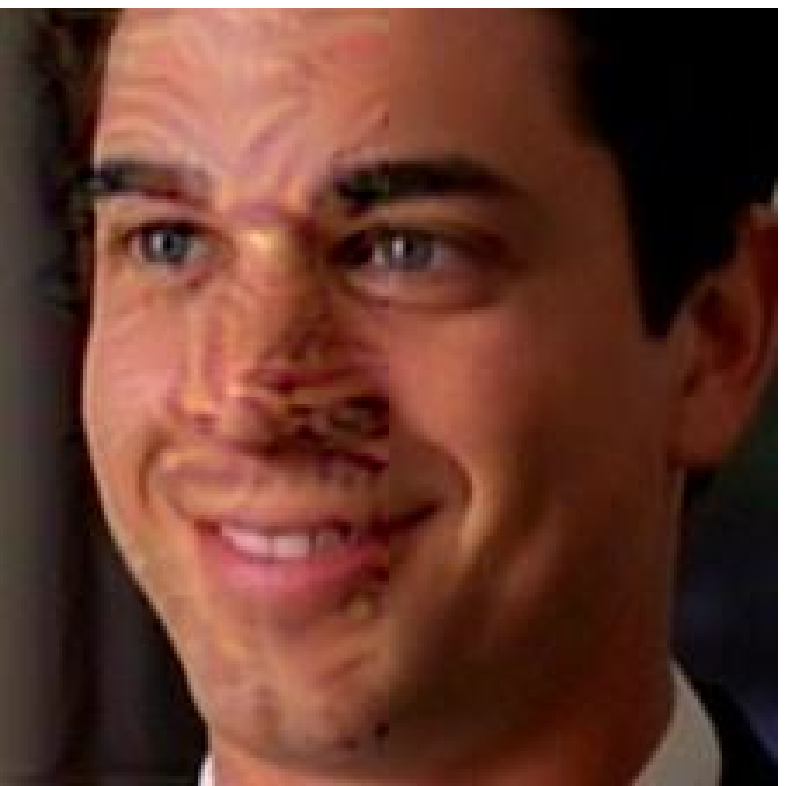}
        \caption{}
    \end{subfigure}%
    ~ 
\begin{subfigure}{0.275\columnwidth}
        \centering
        \includegraphics[height=0.8in]{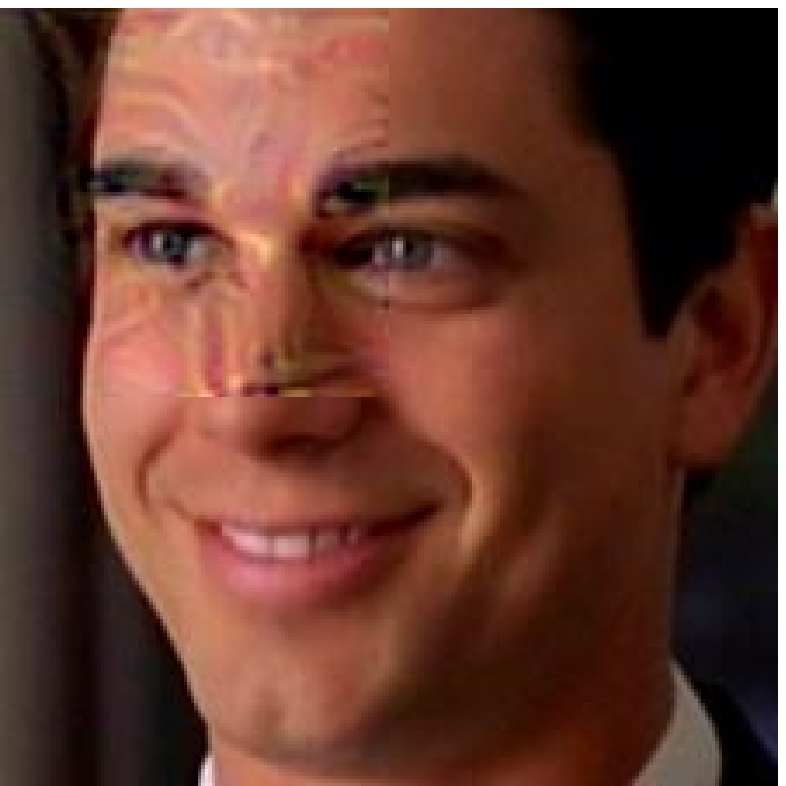}
        \caption{}
    \end{subfigure}%
\caption{Perturbed images with restrictions on the location of pixels to be
  perturbed. In (a), all pixels are to be perturbed, $\sigma=2.7160$. In (b), only left half of
  the image is allowed to be perturbed to achieve the same goal as (a). In (c), only top left quarter is allowed to be perturbed to achieve the same goal as (a).}
\label{fig:halfquarter}
\end{figure}

To ensure that an impersonation attack is imperceptible (i.e., does not raise
suspicion for human observers), the attackers should modify the faces such that
visibility of the modifications is minimal. In this section, we describe the
attack model and how the magnitude of the perturbation are meatured. The lower the magnitude of
the perturbation, the higher the imperceptibility \cite{sharif2016accessorize,
  szegedy2013intriguing}.

\subsection{Attack Model}
We assume that the attacker mounts the impersonation attack after the system has
been trained. This implies that the adversary cannot "poison" the FRS by
altering training data or by injecting mislabeled data. Rather, the adversary
can only alter the composition of input images based on the knowledge of the
underlying DNN model. Our attack model is consistent with IoT  access
control attack scenarios where the attacker cannot tamper with the manufacturing
of the commercial smart devices. In this paper, we mainly focus on a white-box
model in which the attacker knows the DNN architecture and the parameters of the
FRSs being attacked. This is supported by the fact that it is possible to
train local models that can infer the functionality of the target FRSs
\cite{shokri2016membership} and carry transfer attacks to the target FRSs. However, in Section \ref{sec:cross}, we also
examine a black-box model by evaluating how well the perturbed images generated
for one model can be successful in the impersonation attack on another model.

\subsection{Perturbation Vector}
\label{sec:algr}
Suppose the finite set of people's identities
(i.e., labels) to be detected by the FRS is $\Cc$, with $|\Cc| = N$.  Further,
suppose that each input image is given as an RGB vector $\xv$ and the ground
truth label of $\xv$ is given by $c_x \in \{1,2,\cdots,N\}$.

A DNN-based FRS implements a high-dimensional non-linear function which maps an
input $\xv$ to an output probability vector $f(\xv)$ of length $N$, where each
element in the output vector represents the probability that $\xv$ matches the
corresponding label. In addition, the label that corresponds to the largest
entry in $f(\xv)$ is output as the recognition result. Consequently, a correct
recognition result is realized when $c_x$th entry of $f(\xv)$ is the maximum
entry. Thus, the ideal output $f(\cdot)$ is a one-hot vector, i.e., only the
$c_x$th entry has value $1$ and all the other entries are zero.

To impersonate a target $c_t$, the attacker with an input image vector $\xv_a$
thus finds a perturbation vector $\rv$ such that $c_t$th entry of
${f(\xv_a+\rv)}$ is the maximum one.  To measure the error in the output of the
FRS with the adversarial input $\xv_a+\rv$, we adopt the \textit{softmaxloss}
score \cite{parkhi2015deep}. For an input vector $\xv_a$ and a given label
$c_t$, the \textit{softmaxloss} function is defined as:
\begin{equation}
  softmaxloss( f(\xv_a), c_t)=-\log(\frac{e^{<h_{c_t}, f(\xv_a)>}}{\sum_{c=1}^{N}e^{<h_{c}, f(\xv_a)>}}),
\end{equation}
where $<\cdot,\cdot>$ denotes inner product between two vectors and $h_c$ is the
one-hot vector corresponding to label $c$. Note that the value of
\textit{softmaxloss} score is low when the DNN outputs the label as $c_t$ and
high otherwise. The attacker's goal is to achieve a
$softmaxloss(f(\xv_a+\rv),c_t)$ that is low enough such that $c_x$th entry of
$f(\xv)$ is the maximum entry, while minimizing $||\rv||$. In other words, the
attacker solves the following optimization problem.

\begin{equation}
  \rv^* = \argmin_{\rv} {softmaxloss(f(\xv_a+\rv),c_t) + \alpha ||\rv||}.
\label{eq:opt}
\end{equation}

In \eqref{eq:opt}, $\alpha$ is weight factor used to balance impersonation error
and imperceptibility. 
As discussed in \S~\ref{sec:related}, BIM algorithm~\cite{kurakin2016adversarial} as shown in Algorithm~\ref{alg:algr} is used to solve this optimization problem.


\begin{algorithm}
\caption{Computing perturbation vector.}\label{alg:algr}
\begin{algorithmic}[1]
\State \text{Input: image $\xv_a$, target identity $c_t$}
\State \text{Output: impersonation perturbation $\rv$}
\State Initialize $\rv \gets \zerov$
\While {$\xv_a+\rv$ is not recognized as $c_t$}

\State $\Delta \rv = \text{argmin } softmaxloss(f(\xv_a+\rv+\Delta \rv),c_t)$ 
\State Quantize the additional perturbation: 
$\Delta \rv^{\prime} \gets \Delta \rv$
\State Update the perturbation:
$\rv \gets \rv+\Delta \rv^{\prime}$
\EndWhile
\end{algorithmic}
\end{algorithm}

\subsection{Measuring Imperceptibility}
\label{sec:sigma}

Using \eqref{eq:opt}, the attacker can always find perturbation vectors that allow
desired misclassification of input vectors. However, the produced attack image,
i.e., $\xv_a + \rv^*$ is not guaranteed to be ``imperceptible" to humans. In
other words, the perturbation vector could be too large.  This would cause the
produced perturbed image to be quite distinguishable from the original attacker
image.  To quantify the effectiveness of the attack in various settings, we
measure per pixel per color channel magnitude of perturbation using the root
mean square error (RMSE) between the original and perturbed images. In
particular, suppose that the images are of width $W$, height $H$ and number of
color channels $D$. Let the total number of dimensions in an image vector be
$M = W \times H \times D$. Given an input and perturbed image vectors
$\xv, \xv' \in \left\{0,1,\cdots, 255\right\}^{M}$, the RMSE (we also use the
term ``noise level'') is given by the following.
\begin{equation}
  \sigma(\xv,\xv')=\sqrt{\frac{1}{M}\sum_{i=1}^{M}(x(i)-x'(i))^2},
\end{equation}
where $x(i)$ is the $i$th component of $\xv$, and $\sigma$ is in pixel-value
units, where $\sigma\in[0,255]$.

To get a sense of what values of $\sigma$ renders a perturbed attack image easy
to identify, we show images of an attacker with varying levels of perturbation
in Fig. \ref{fig:perturbedImg}. We note that for $\sigma>2$, it is easy to
identify the noisy pixels in the perturbed images.

\subsection{Physical Imperceptibility}

If the attackers want to realize this perturbation
physically (via using various paraphernalia such as dummy
faces, or 3D-printed glasses), the amount of perturbation will need to be limited in terms of
either (a) the maximum number of pixels to which the noise is added, or (b) the
locations of those pixels \cite{sharif2016accessorize}, or (c) both. In
Fig. \ref{fig:halfquarter}, we study the effects of such limitations. 
We fix the attacker image and a target label, and then find the adversarial
images when the entire image can be perturbed, as well as when only the left half
and top left quarters of the image pixels are to be perturbed. As shown in the
figure, the noise level increases significantly and the pattern is
perceptible. Thus, one can expect the attack to be much harder in these
cases. In the rest of the paper, we only consider scenarios in which the full
attacker image is subject to perturbation. This reflects a worst case scenario
analysis from the defender's perspective. Even in this scenario, we show that it
can be hard for an attacker to launch the attack in all possible scenarios.


%% file: exp3.tex
\section{Experiments}
\label{sec:experiments}

\begin{figure*}[t]
    \centering
    \begin{subfigure}[t]{0.32\textwidth}
        \centering
        \includegraphics[height=0.8in]{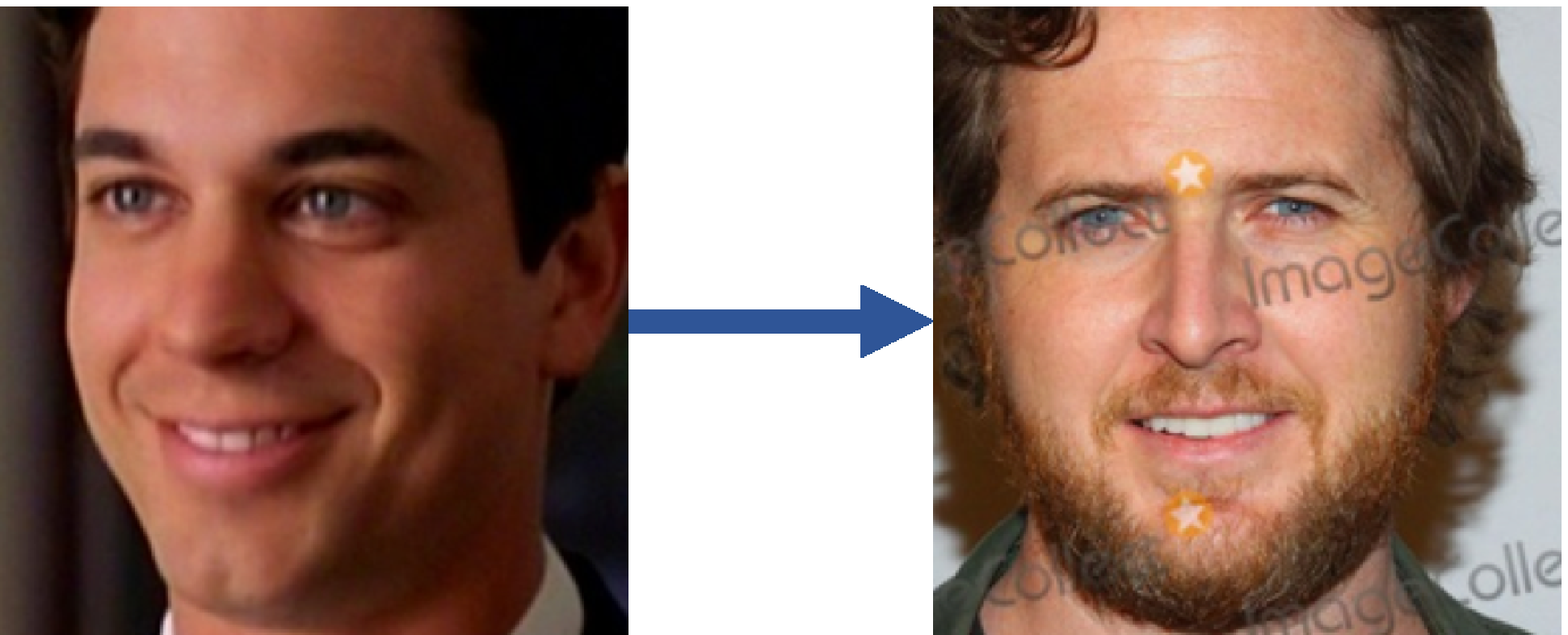}
        \caption{Impersonating A.J. Buckley, $\sigma=0.88$}
    \end{subfigure}%
    ~
    \begin{subfigure}[t]{0.32\textwidth}
        \centering
        \includegraphics[height=0.8in]{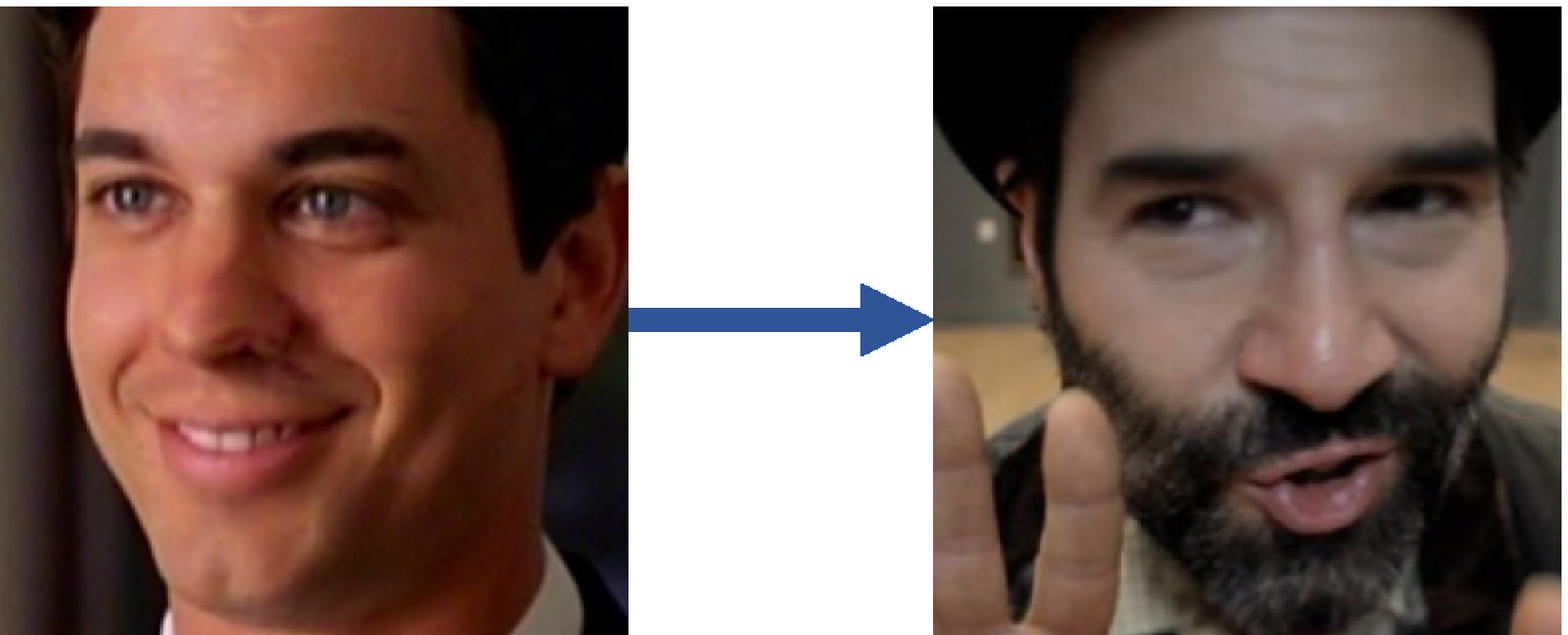}
        \caption{Impersonating Adam Buxton, $\sigma=1.65$}
    \end{subfigure}%
    ~ 
   \begin{subfigure}[t]{0.32\textwidth}
        \centering
        \includegraphics[height=0.8in]{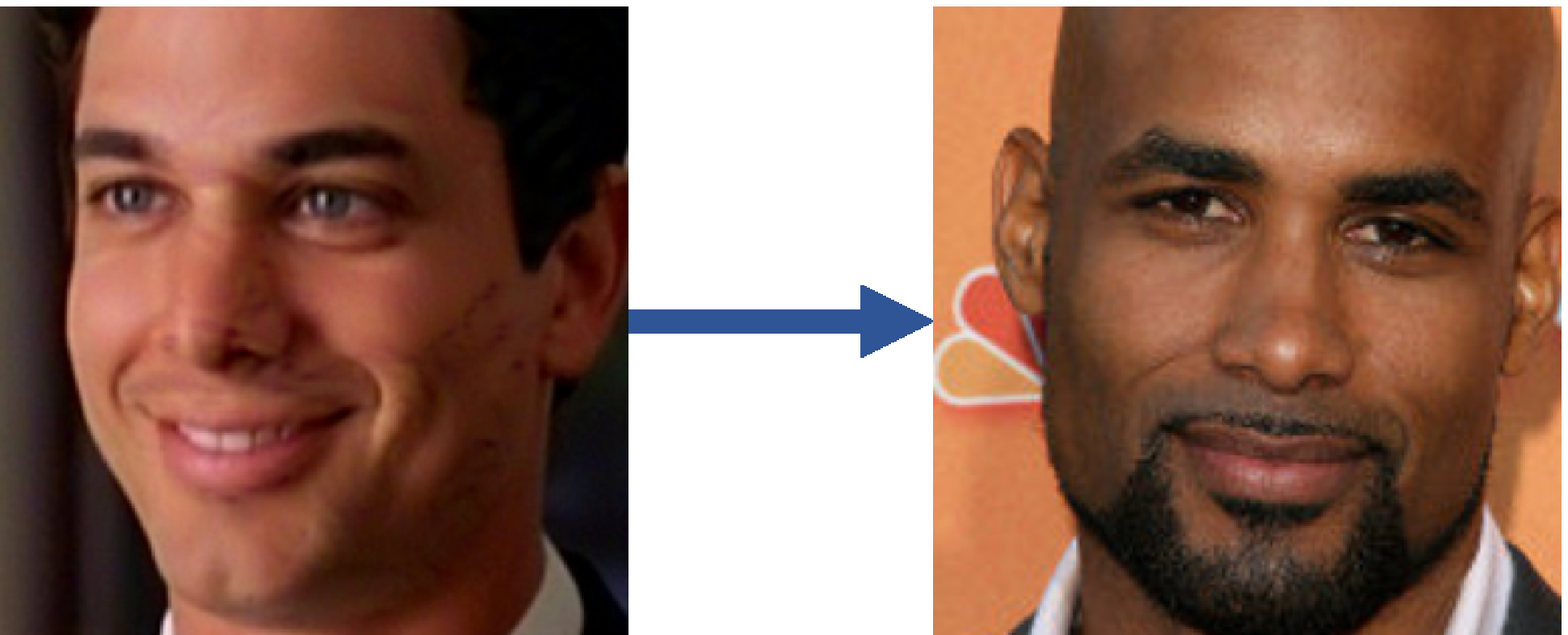}
        \caption{Impersonating Boris Kodjoe, $\sigma=2.26$}
    \end{subfigure}%
    \caption{Different noise levels needed for Micheal Crichton to impersonate three different identities. It is rather easy for Micheal Crichton to impersonate A.J. Buckley; and hard to impersonate Boris Kodjoe } \label{fig:DiffGen}
\end{figure*}

\begin{figure*}
\centering
\includegraphics[width=1.7\columnwidth]{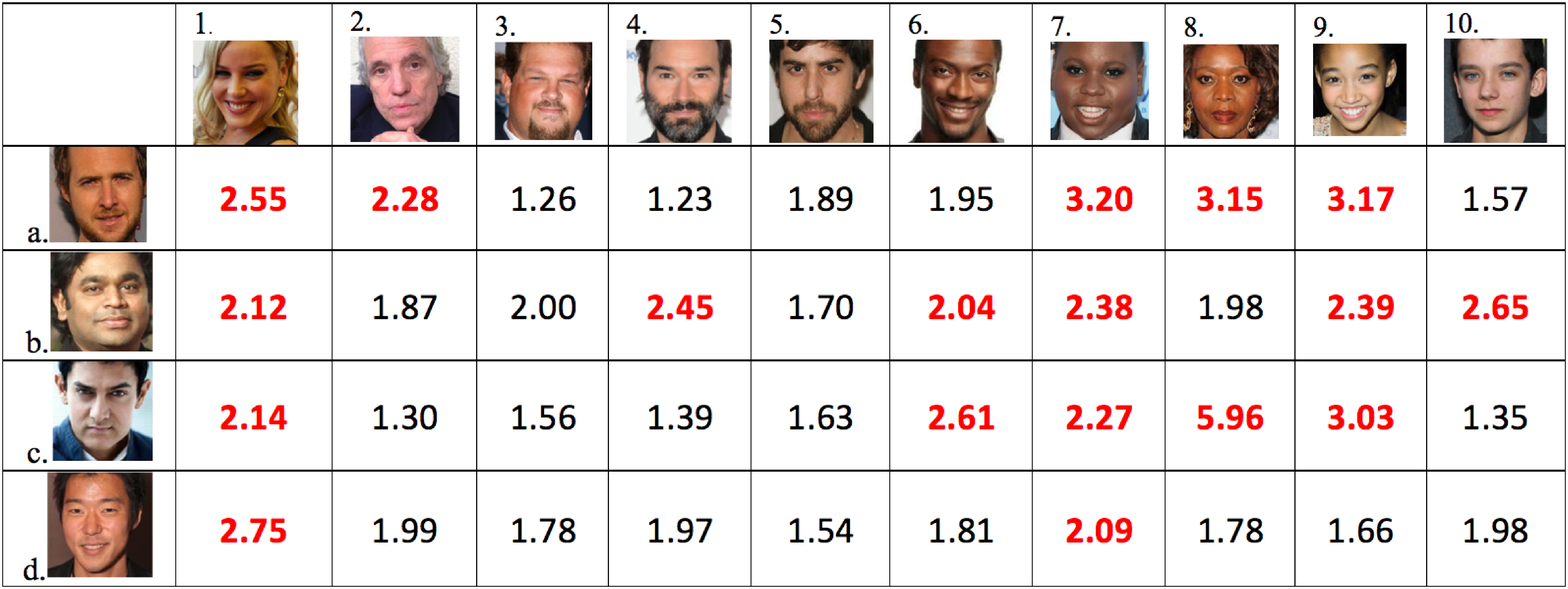}
\caption{Noise level $\sigma$ required for an attacker (a-d) to impersonate a
  target (1-10). It is easier for the considered attackers (who are all male with pale skin
color) to impersonate targets who are also male with pale skin color, as
compared to impersonating other targets. The noise levels needed to
impersonate target 1 are all large. Impersonating targets 6-10 seems to require
larger noise levels. } \label{fig:generaltable}
\end{figure*}

In this section, we detail the results of our measurement study towards getting an
in depth understanding of the practicality of imperceptible
impersonation attacks on DNN-based FRS and the factors that influence such attacks.
\subsection{Experimental Setup}
\label{sec:setup}
The FRS used in our experiments is VGG-Face~\cite{parkhi2015deep}, one of the most well-known and highly accurate
face recognition systems as discussed in \S~\ref{sec:related}. 
The analysis is based on the VGG-Face
dataset \cite{parkhi2015deep}, which contains $N=2622$ identities of
celebrities, and approximately $1000$ facial images per identity; this
translates to a total of about $2.6$ million images.

\subsection{Case studies}
\label{sec:general}

To begin with, we use the face image of Micheal Crichton as the attacking image,{\color{black} (i.e., the input)} and study whether some targeted individuals are harder than others to impersonate with the attacking image. Fig.~\ref{fig:DiffGen} shows the minimum perturbations needed for the attacking image to impersonate three different individuals. We observe that it is rather easy for Micheal Crichton to impersonate A.J. Buckley. However, when it comes to impersonating Boris Kodjoe, the perturbation gets larger and is noticeable by human.

For a more general case study, Fig. \ref{fig:generaltable} shows the noise level
$\sigma$ needed to achieve a successful attack by each attacker depicted on the
column, to impersonate each target depicted on the row. It is clear that,
different attackers need different values of $\sigma$ to successfully
impersonate different targets. Interestingly, the patterns of large
perturbations (marked in red) seen in Fig. \ref{fig:generaltable} suggest that
it is easier for the considered attackers (e.g., who are all male with pale skin
color) to impersonate targets who are also male with pale skin color, as
compared to impersonating other targets. In addition, the noise levels needed to
impersonate target 1 are all large, which is possibly due to a difference in
gender. Furthermore, we see that impersonating targets 6-10 seems to require
larger noise levels. This can be attributed to differences in skin color, age,
or a combination of both. This motivates our study to further examine the impact
of these factors in Section \S~\ref{sec:groups}.

\begin{figure*}[t!]
\begin{center}
\parbox{3.2in} {
\centerline{\includegraphics[width=\columnwidth]{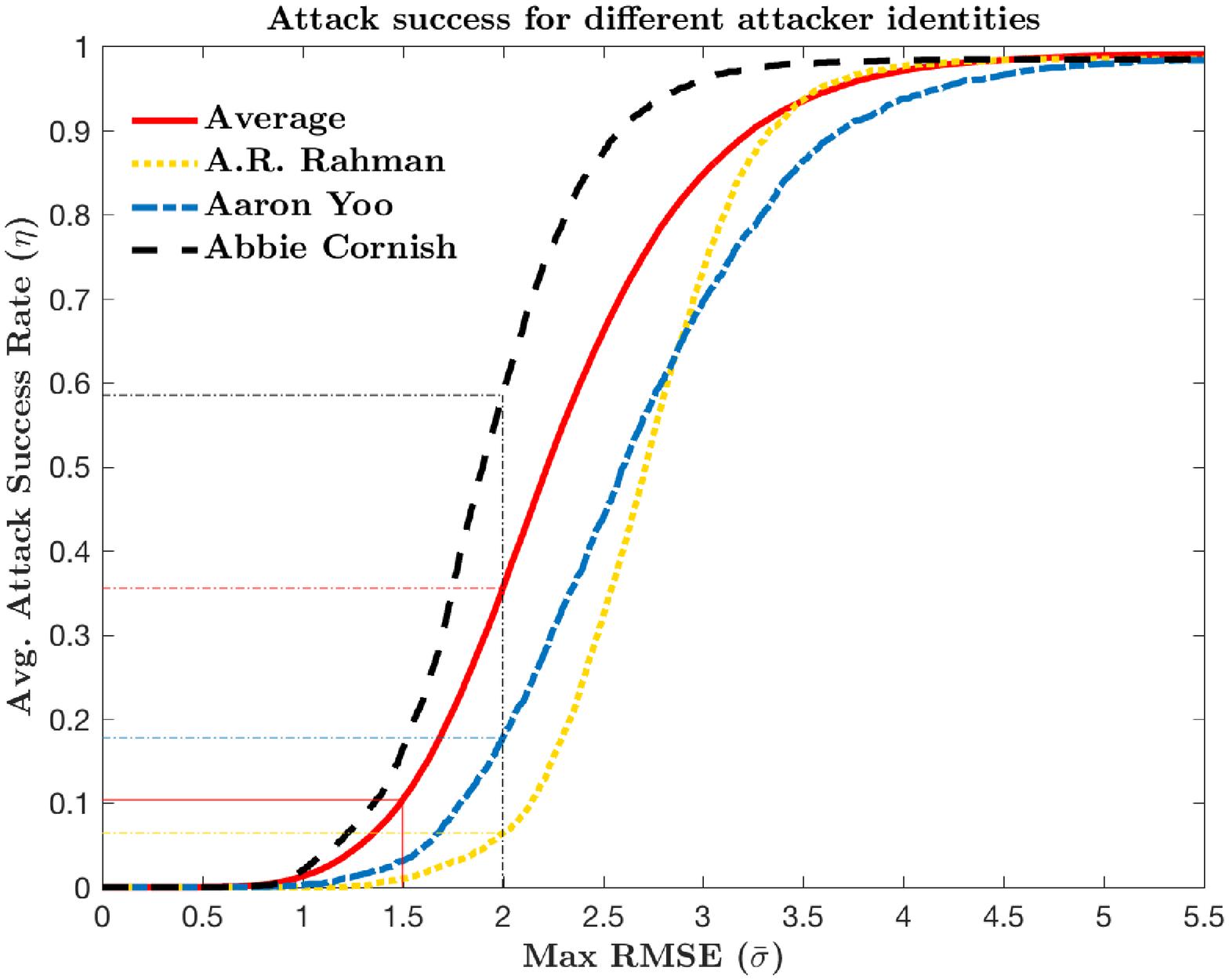}}
\caption{Impersonation attack performance. Abbie Cornish (female, white, young) can more successfully
impersonate others, on average, compared to A.R. Rahman (male, Indian, young)
and Aaron Yoo (male, Asian, young).} \label{fig:generalfig}}
\makebox[.08in] {}
\parbox{3.2in} {
\centerline{ \includegraphics[width=\columnwidth]{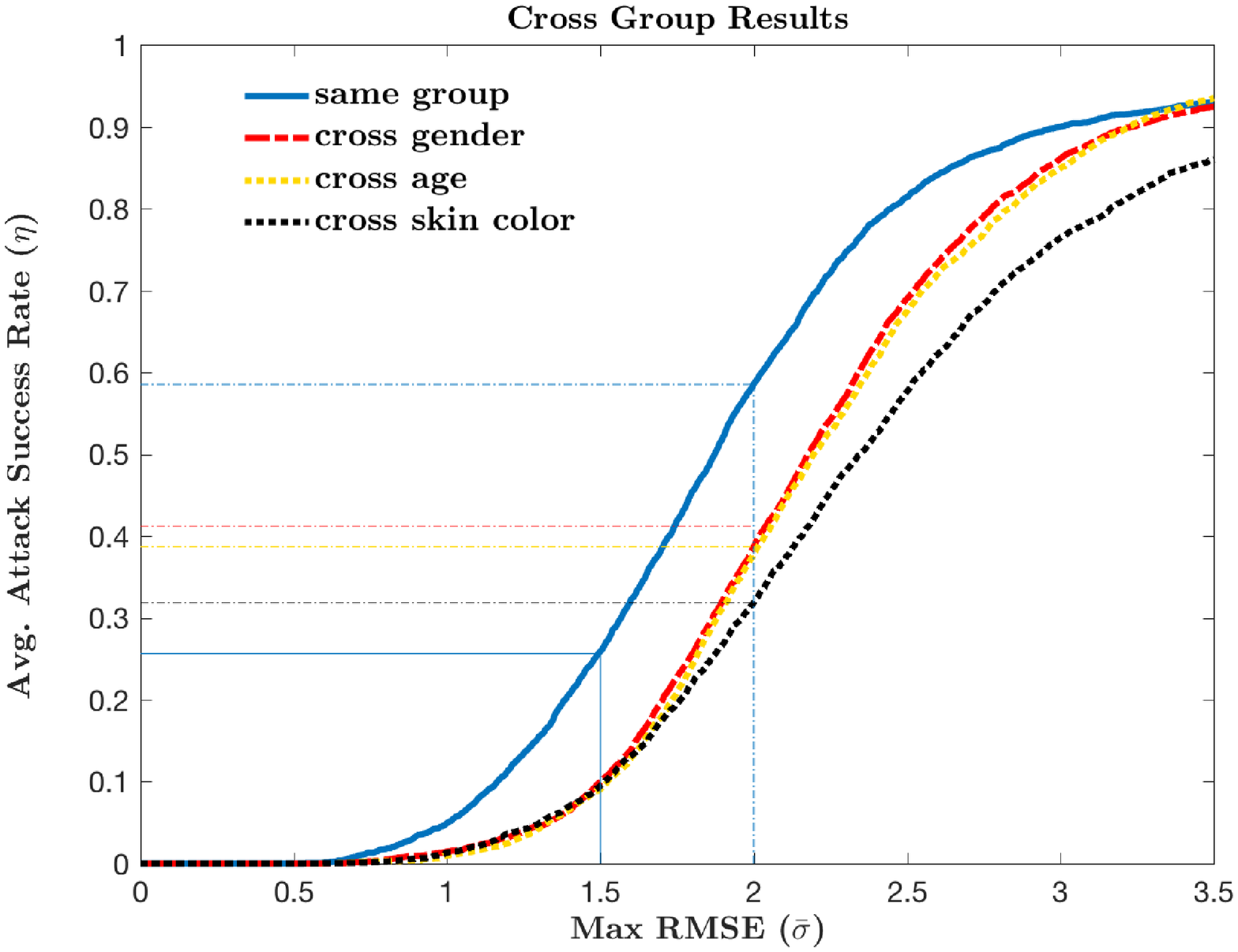}}
\caption{Cross group impersonation attack performance. It is easier for an attacker to
impersonate a target identity having the same attributes (gender, skin color,
age). Impersonation across different skin color is the most hardest.}\label{fig:groups}}
\end{center}
\end{figure*}

Having performed the above preliminary studies, we next look at the statistical
distribution of the ability of an attacker to impersonate different targets,
subject to a constraint on the noise level $\sigma\leq \bar{\sigma}$. We define
the attack success rate $\eta(\bar{\sigma})$ as the percentage of target labels
which an attacker can impersonate for a given $\bar{\sigma}$. In
Fig. \ref{fig:generalfig}, we show the success rates $\eta$ for three different
attackers impersonating all other remaining labels in the VGG-Face dataset. One
can see that Abbie Cornish (female, white, young) can more successfully
impersonate others, on average, compared to A.R. Rahman (male, Indian, young)
and Aaron Yoo (male, Asian, young). For example, with the threshold
$\bar{\sigma}= 2$, Abbie Cornish can successfully impersonate $58\%$ of all the
labels while A.R. Rahman achieves a success rate of only $6.5\%$ and Aaron Yoo
achieves a success rate of $17.7\%$. This could be attributed to the fact that
the VGG-Face dataset contains more white people than people of other races. We
observe that the gender distribution is almost balanced in the dataset.

To get aggregate results, we randomly sample the VGG-Face dataset to get a
$100$-identity subset $\Sc$. We fix each identity in $\Sc$ as a specific attacker,
and then find the perturbation vector with each of the remaining labels in $\Sc$
as targets, and we compute $\eta(\bar{\sigma})$ for each attacker for a range of
values of $\bar{\sigma}$. We then repeat this experiment $10$ times and compute
the average attack success rate across attackers in all the samples. The results
show that, on average, it is not easy for an attacker to impersonate {\em any}
target identity. In particular, with $\bar{\sigma}=1.5$, the success rate is
only $10.4\%$. We take a deeper look into how the success rate breaks down
within different groups of people in the following section.

\subsection{Factors that influence the attack}
\label{sec:groups}

Next, we take a closer look at the extent to which various factors, discussed in
\S~\ref{sec:general}, influence an attacker's ability to carry out
an imperceptible impersonation attack. Specifically, we consider different
groups of identities based on gender, skin color, and age attributes. We manually
label the dataset to produce four groups: (a) white young male (100 identities),
(b) white young female (100 identities), (c) black young male (69 identities),
and (d) white old male (100 identities). We do not consider other attribute
combinations, such as black young female, or white old female, because the
majority of the images in the VGG-Face dataset are for white skin color, and
young people. For group (c), we only have 69 identities due to limitedness
of data points matching such attributes. To reduce errors in labeling, each of
the authors of the paper manually labeled the dataset independently and we
considered only the images with unanimously common labels in our group dataset.
In addition, when labeling, we discard an identity whenever its attributes are
hard to label manually.

 To investigate the impact of the aforementioned attributes on the
 imperceptible impersonation attack,
we conduct four experiments based on the four groups:
\squishlist
\item Take people in group (a) as attackers trying to impersonate the other
  people in group (a); this case reflects {\it same group} impersonation
  measurements;
\item take people in group (a) as attackers trying to impersonate people in
  group (b); this case represents {\it cross gender} impersonation measurements;
\item take people in group (a) as attackers trying to impersonate people in
  group (c); this case reflects {\it cross skin color} impersonation
  measurements;
\item take people in group (a) as attackers trying to impersonate people in
  group (d); this case counts for {\it cross age} impersonation measurements.
\squishend

In Fig. \ref{fig:groups}, we plot the average attack success rate versus
different perturbation constraint $\bar{\sigma}$, for each of the four
aforementioned experiments. We note that it is easier for an attacker to
impersonate a target identity having the same attributes (gender, skin color,
age). For the same group experiment, with the threshold $\tilde{\sigma} = 1.5$,
the success rate is $25.65\%$. Recall that the aggregate success rate in
\S~\ref{sec:general} is only $10.4\%$. Moreover, as shown in the figure, it is
relatively easier for an attacker to impersonate a target with a different
gender or age than to impersonate a target with different skin color. For
example, with the threshold $\tilde{\sigma}= 2$, the success rate for cross skin
color is only $31.85\%$ while the success rate for cross age and gender are
around $40\%$. These results seem consistent with (and can be explained by)
observations that have been previously reported in computer vision
literature~\cite{choi2009color, torres1999importance, shih2005comparative,
  karimi2006comparative}. Specifically, these papers show that in several
scenarios, shape and texture cues suffer from degradation (affecting age or
gender) and the color feature becomes dominant \cite{yip2002role}. Thus, we
conclude that the VGG-Face model relies less on features such as shape and
texture as compared to color.

\subsection{Universal Perturbation Results}
\label{sec:universal}

In a realistic setting, an attacker may want one universal perturbation to impersonate the target identity for all the face images captured in different settings  
such as pose, camera angle, and lighting
conditions. In order to launch the impersonation attack in the presence of these
variations, an attacker will need to find a single perturbation vector
$\tilde{\rv}$ that allows misclassification of a set of his/her own images
$\Xc_a$ of size $K$, to the target victim label, thus accounting for as many
conditions as possible. In other words, the attacker needs to construct a vector
$\tilde{\rv}$ such that $f(\xv_a + \tilde{\rv})=c_t, \, \forall \xv_a \in\Xc_a$
for some given target label $c_t$.

The approach for calculating $\tilde{\rv}$ is similar to the one described in
\S~\ref{sec:algr}. The only difference is that now the objective function
changes to the following.

\begin{equation}
  \tilde{\rv} = \argmin_{\rv} {\sum_{\xv_a\in\Xc_a}softmaxloss(f(\xv_a+\rv),c_t) + \alpha ||\rv||}.
\label{eq:opt2}
\end{equation}

The condition to stop the iterations now requires that all $K$ images to be
misclassified as the target label, upon adding on the same perturbation vector.

\begin{figure}[t]
    \centering
    \begin{subfigure}[t]{0.275\columnwidth}
        \centering
        \includegraphics[height=0.85in]{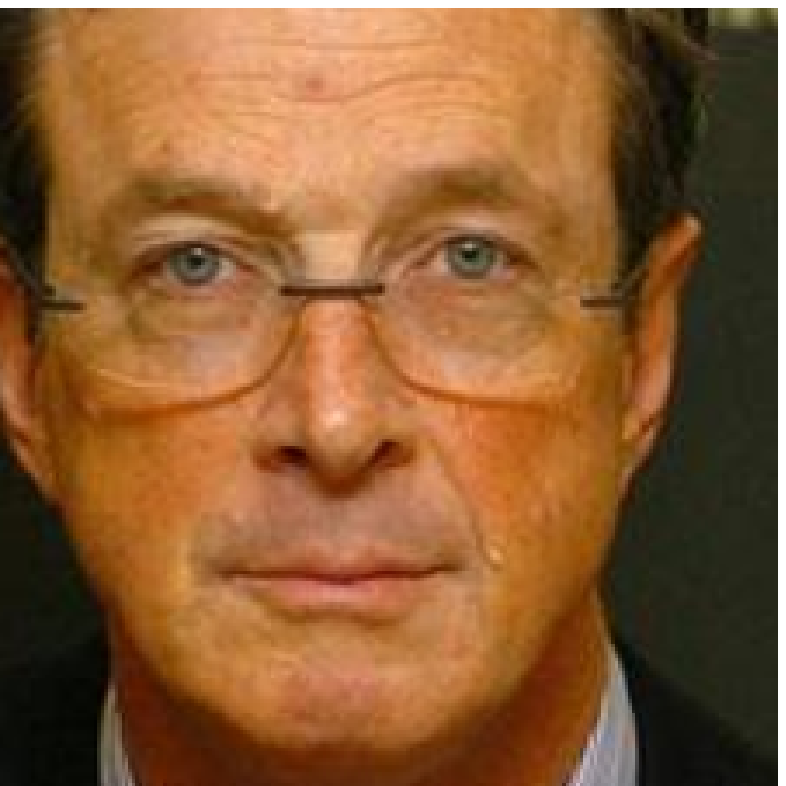}
    \end{subfigure}%
    ~ 
    \begin{subfigure}[t]{0.275\columnwidth}
        \centering
        \includegraphics[height=0.85in]{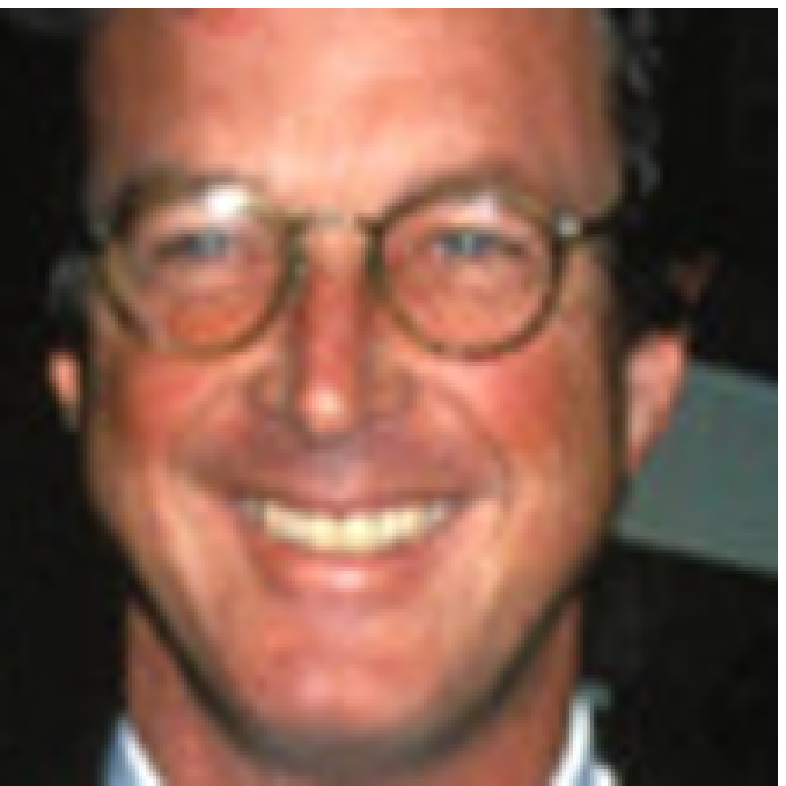}
    \end{subfigure}%
    ~
    \begin{subfigure}[t]{0.275\columnwidth}
        \centering
        \includegraphics[height=0.85in]{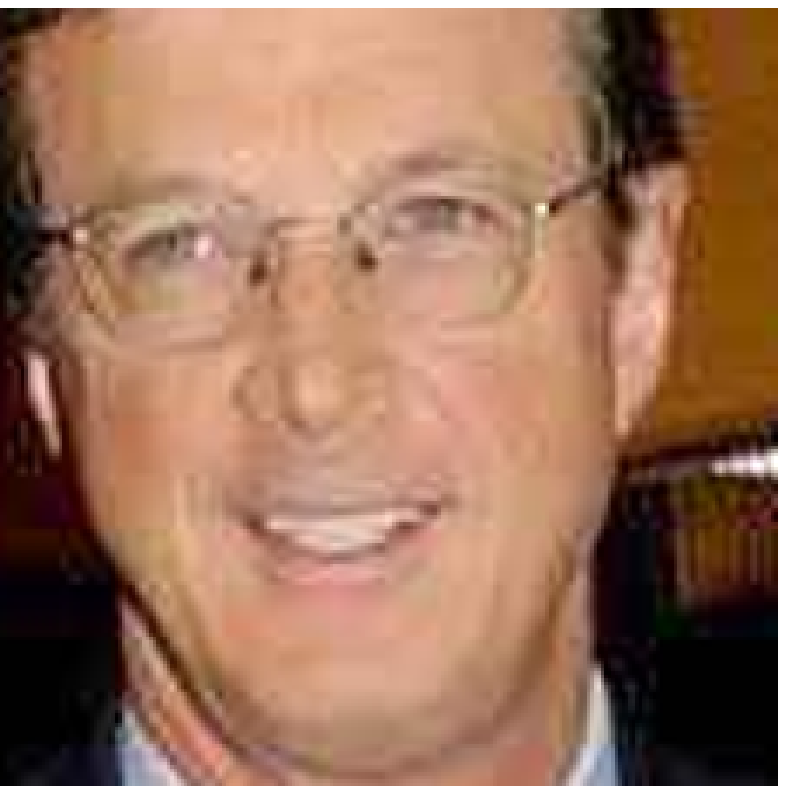}
    \end{subfigure}%
  \caption{A set of face images of Micheal Crichton. $\Xc=\{\xv_1,\xv_2,\xv_3\}$ }
  \label{fig:ThreeIms}
  \end{figure}

\begin{figure}[t]
 \centering
   \begin{subfigure}[t]{0.275\columnwidth}
        \centering
        \includegraphics[height=0.85in]{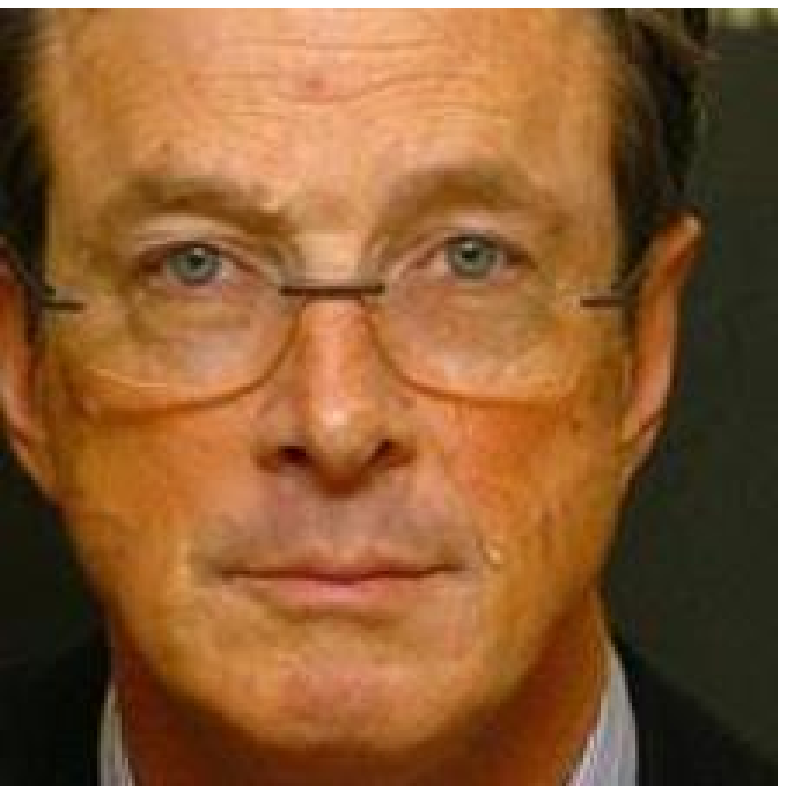}
    \end{subfigure}%
    ~
 \begin{subfigure}[t]{0.275\columnwidth}
        \centering
        \includegraphics[height=0.85in]{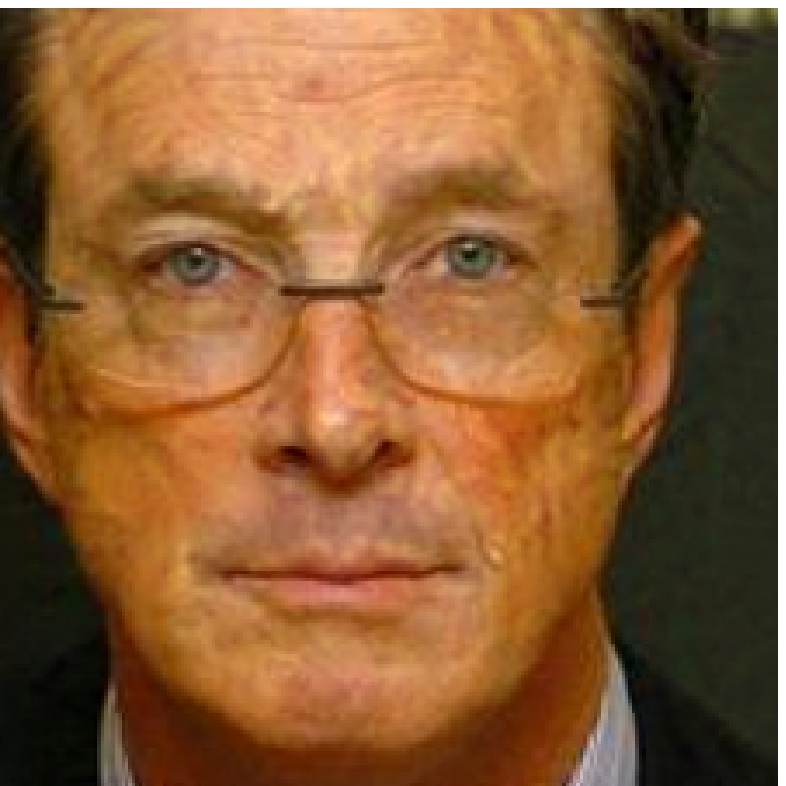}
    \end{subfigure}%
    ~ 
\begin{subfigure}[t]{0.275\columnwidth}
        \centering
        \includegraphics[height=0.85in]{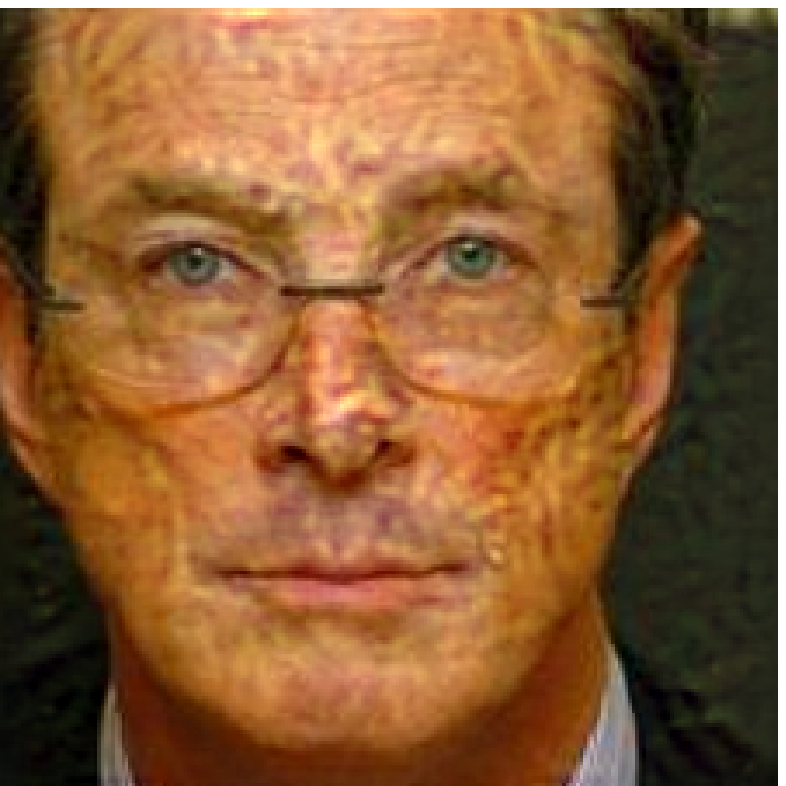}
    \end{subfigure}%
    \caption{Universal perturbations visualization. Three perturbations are universal to different number of attacking images. Left: universal for $\{\xv_1\}$, $\sigma=1.7509$; Middle: universal for $\{\xv_1,\xv_2\}$, $\sigma=3.6027$; Right: universal for $\{\xv_1,\xv_2,\xv_3\}$, $\sigma=7.8877$.
    The perturbations are more perceptible as more attacking images are considered in computing the universal perturbation vector.} 
    \label{fig:ThreeUP}
\end{figure}

In Fig.~\ref{fig:ThreeIms}, we show an example of an attacker with three
images, $\Xc_a=\{\xv_1, \xv_2, \xv_3\}$. In Fig.~\ref{fig:ThreeUP}, we show the output
perturbed image $\xv_1+\tilde{\rv}$ when $\tilde{\rv}$ is computed using only
image $\{\xv_1\}$, images $\{\xv_1, \xv_2\}$, and all the images $\{\xv_1, \xv_2, \xv_3\}$, respectively. It is
evident that the attacker image is more perceptible as more attack images are
considered in computing the universal perturbation vector $\tilde{\rv}$.

\begin{figure}
\centering
\includegraphics[width=\columnwidth]{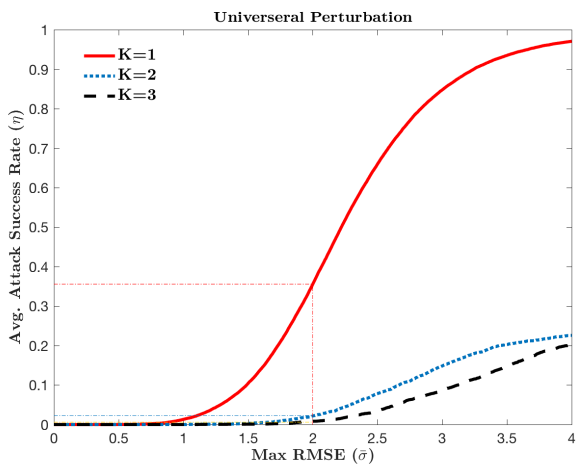}
\caption{Universal perturbation impersonation attack
  performance. K is the number of attacking images to generate the universal perturbations. The attackers' ability to
impersonate given targets is significantly reduced when the perturbations are required to be universal to multiple attacking images.} \label{fig:universal}
\end{figure}

In Fig.\ref{fig:universal}, we plot the average success rate for an attacker
employing universal perturbation. Here, we randomly sample $100$ identities from
the VGG-Face dataset and let them impersonate each other. We conduct this experiment
$10$ times and average the results. The results show that the success rate is
strictly decreasing when a universal perturbation vector is required to perturb
multiple attacker images. More importantly, the attackers' ability to
impersonate a given target is significantly reduced with even slight increases
in $K$. For example, the success rate with threshold $\tilde{\sigma}=2$ is
$39.9\%$ for $K =1$ (the case considered in \S~\ref{sec:general} and
\S~\ref{sec:groups}). However, when we increase $K$ to $2$, the success rate
drops dramatically to $2.28\%$ and the success rate when $K=3$ is only $0.6\%$,
which suggests that the impersonations can almost fail all the time, if the
attacker seeks to be imperceptible.

\subsection{Cross Model Measurements}
\label{sec:cross}
Recently, it has been shown that adversary examples that are successfully misclassified by one
trained DNN model can also cause misclassifications in other (different) DNN models that
have different hyperparameters \cite{szegedy2013intriguing, moosavi2016universal}. However, it is unclear whether different models could misclassify the perturbed images to the same target classes, which is the key characteristic for 
determining if white-box impersonation attacks can easily extend to black-box attacks. 
To check whether our perturbed images targeting impersonation generalize across different
DNN models, we fine-tune the AlexNet DNN \cite{krizhevsky2012imagenet} on the
VGG-Face dataset, and test the impersonation attack success ratio on the AlexNet
model using the perturbed images generated using VGG-Face model.

We test $10,000$ perturbed images generated by VGG-Face, and find that {\it
  none} are classified as the victims by the AlexNet, but most of them are misclassified by the AlexNet. This
significant result indicates that impersonation attacks do not easily transfer across different DNN models. It will be extremely hard for the attacker to use
the perturbation vectors to fool a DNN model different from the one used to
generate them. Thus cross model validation could significantly enhance the
robustness of face recognition based access control in IoT systems.

\subsection{Detecting and removing perturbations}
\label{sec:noise}


Finally, we test whether de-noising~\cite{patidar2010image} (which could be used
by an IoT access control system) affects the potency of the attack. Three
standard de-noising filters are considered in our experiments: average filter,
median filter, and Wiener filter. We test $100$ different perturbed images,
and find that all of them are still misclassified as the targets. This suggests
that de-noising does not hurt the attack. This is because de-noising filters
assume a certain pattern of noise, which is unlikely to be what is used by the
attacker for generating the perturbations.

We conclude that traditional noise detection and de-noising algorithms are not
helpful in countering the imperceptible impersonation attack since the
perturbation generated is structured.

\subsection{Summary of results}
Below is a summary of our take-aways based  on the results in \S~\ref{sec:general} to \S~\ref{sec:noise}. 
{\bf (a)} DNNs are vulnerable to adversary examples. However, in contrast to recent
  work in the literature, we find that the average success rates of the
  imperceptible impersonation attack are low.
{\bf (b)} Attackers can achieve better success rates by choosing targets with similar
  attributes; in particular choosing targets with same skin color helps.
{\bf (c)} When variations, such as pose, camera angle and lighting conditions are
  considered, the attack is significantly less successful.
{\bf (d)} Perturbed images do not generalize well across different DNN models.
{\bf (e)} Current noise estimation and de-noising methods do not  adversely impact the imperceptible impersonation attack.


%% file: conc2.tex
\section{Conclusion}
\label{sec:conclusion}
The security of face recognition is an important toptic as face recognition is more and more used in IoT access
control.
In this paper, we perform an in-depth measurement study of the generality and
efficacy of imperceptible impersonation attacks that have recently gained
popularity.
Our study is done using a very large dataset. We find that it is hard
for a given adversary to impersonate an arbitrary target victim without making
perceptible changes to her face. Further, we show that several factors such as
age, race and gender of the attacker and victim influence the efficacy of the
attack and we quantify the impact of each. We also show that, in a realistic
scenario where the attacker seeks to be robust to different poses or variations
in environmental conditions, the attack becomes more difficult or even
impossible. Based on this, we suggest the use of cross-model verifications as
well as multi-views, which can potentially counter such attacks very
effectively.
\section*{Acknowledgments}
{
This research was partially sponsored by the U.S. Army Combat Capabilities Development Command Army Research Laboratory and was accomplished under Cooperative Agreement Number W911NF-13-2-0045 (ARL Cyber Security CRA). The views and conclusions contained in this document are those of the authors and should not be interpreted as representing the official policies, either expressed or implied, of the Combat Capabilities Development Command Army Research Laboratory or the U.S. Government. The U.S. Government is authorized to reproduce and distribute reprints for Government purposes notwithstanding any copyright notation here on.
}